\documentclass[times,twocolumn,final,nopreprintline]{elsarticle}

\usepackage[utf8]{inputenc}

\usepackage{framed,multirow}

\usepackage{amssymb}
\usepackage{latexsym}

\usepackage{url}
\usepackage{xcolor}

\usepackage{hyperref}

\definecolor{newcolor}{rgb}{.8,.349,.1}

\usepackage{amsmath,amsfonts,amssymb,amsthm}
\usepackage{mathtools}
\usepackage{commath}

\journal{Medical Image Analysis}

\begin{document}
	

	\title{Combined tract segmentation and orientation mapping for \newline bundle-specific tractography}%
	
	\author[1,2]{Jakob Wasserthal}
	\author[1]{Peter F. Neher}
	\author[4]{Dusan Hirjak}
	\author[1,3]{Klaus H. Maier-Hein\corref{cor3}}
	
	\cortext[cor3]{Corresponding author: 
		Tel.: +49-6221-42-3545;}
	\ead{k.maier-hein@dkfz.de}
	
	\address[1]{Division of Medical Image Computing (MIC), German Cancer Research Center (DKFZ), Heidelberg, Germany}
	\address[2]{Medical Faculty Heidelberg, University of Heidelberg, Heidelberg, Germany}
	\address[3]{Section of Automated Image Analysis, Heidelberg University Hospital, Heidelberg, Germany}
    \address[4]{Department of Psychiatry and Psychotherapy, Central Institute of Mental Health, Medical Faculty Mannheim, Heidelberg University, Mannheim, Germany}
	

	\begin{abstract}
		While the major white matter tracts are of great interest to numerous studies in neuroscience and medicine, their manual dissection in larger cohorts from diffusion MRI tractograms is time-consuming, requires expert knowledge and is hard to reproduce. 
		In previous work we presented tract orientation mapping (TOM) as a novel concept for bundle-specific tractography. It is based on a learned mapping from the original fiber orientation distribution function (FOD) peaks to tract specific peaks, called tract orientation maps. Each tract orientation map represents the voxel-wise principal orientation of one tract.
		Here, we present an extension of this approach that combines TOM with accurate segmentations of the tract outline and its start and end region. We also introduce a custom probabilistic tracking algorithm that samples from a Gaussian distribution with fixed standard deviation centered on each peak thus enabling more complete trackings on the tract orientation maps than deterministic tracking. These extensions enable the automatic creation of bundle-specific tractograms with previously unseen accuracy.
		
		We show for 72 different bundles on high quality, low quality and phantom data that our approach runs faster and produces more accurate bundle-specific tractograms than 7 state of the art benchmark methods while avoiding cumbersome processing steps like whole brain tractography, non-linear registration, clustering or manual dissection. Moreover, we show on 17 datasets that our approach generalizes well to datasets acquired with different scanners and settings as well as with pathologies.
		The code of our method is openly available at https://github.com/MIC-DKFZ/TractSeg.
		
	\end{abstract}
	
	\maketitle              

	{\tiny This is the accepted manuscript for https://doi.org/10.1016/j.media.2019.101559.}

	
	\section{Introduction}
	
	\begin{figure*}[!t]
		\centering
		\includegraphics[width=.8\textwidth]{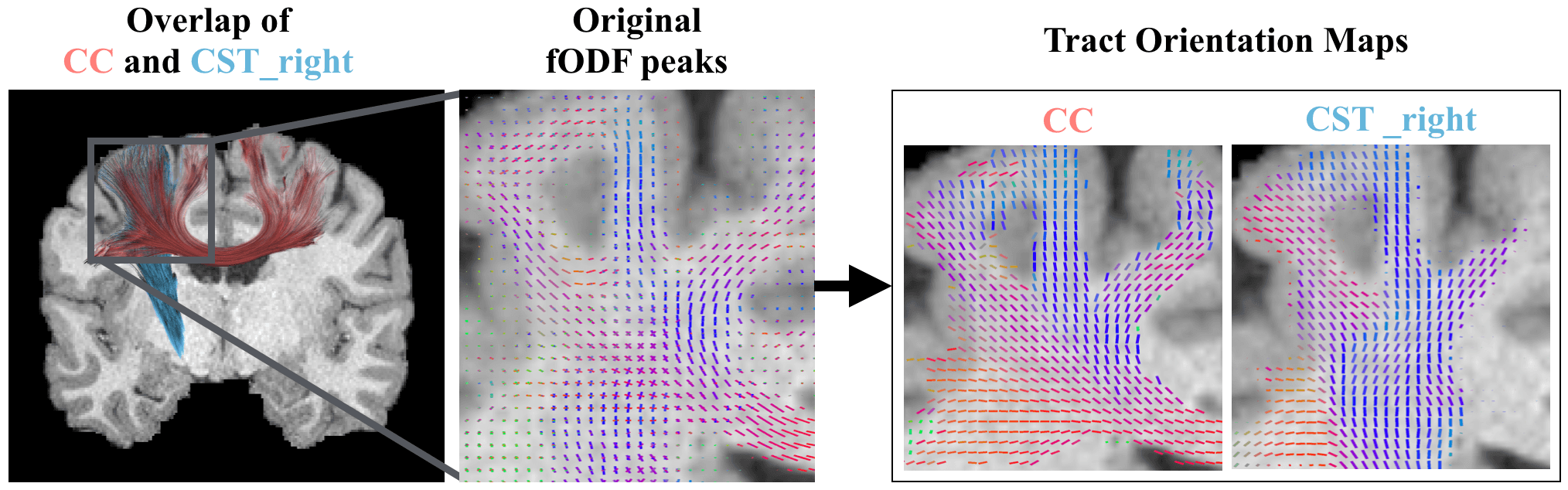}
		\caption{Exemplary depiction of a slice through two of the reference tracts, the original FOD peak image and the corresponding reference TOMs (CST\_right: corticospinal tract; CC: corpus callosum).}
		\label{fig:peaks}
	\end{figure*}
	
	The white matter of the human brain is made up of a large number of individual fiber tracts. Those tracts overlap, resulting in multiple fiber orientation distribution function (FOD) peaks per voxel and larger bottleneck situations with tracts per voxel outnumbering the peaks per voxel. In consequence, tractography is highly susceptible to false positives \citep{maier-hein_challenge_2017,knosche_validation_2015}. The only safe solution around false positives so far is the explicit dissection of anatomically well-known tracts. While manual dissection protocols \citep{stieltjes_diffusion_2013} can be considered the current gold standard, a variety of approaches has already been developed for automating the process: \emph{Region-of-interest-based} approaches filter streamlines based on their spatial relation to cortical or other anatomically defined regions, which are typically transferred to subject space via atlas registration and segmentation techniques \citep{wassermann_white_2016,yendiki_automated_2011}. \emph{Clustering-based} approaches group and select streamlines by measuring intra- and inter-subject streamline similarities, referring to existing reference tracts in atlas space \citep{garyfallidis_recognition_2017,odonnell_automatic_2007,odonnell_automated_2016}.
	Concept-wise, many previous approaches have opted for performing a rather blind whole brain tractography and then investing the effort in streamline space, clearing the tractograms from spurious streamlines and grouping the remaining ones. These approaches often have long runtimes, need several processing steps which are tedious to set up, depend on registration which is error prone and have accuracy which is decent but still leaves room for improvement.
	
\begin{figure*}[!t]
		\centering
		\includegraphics[width=.8\textwidth]{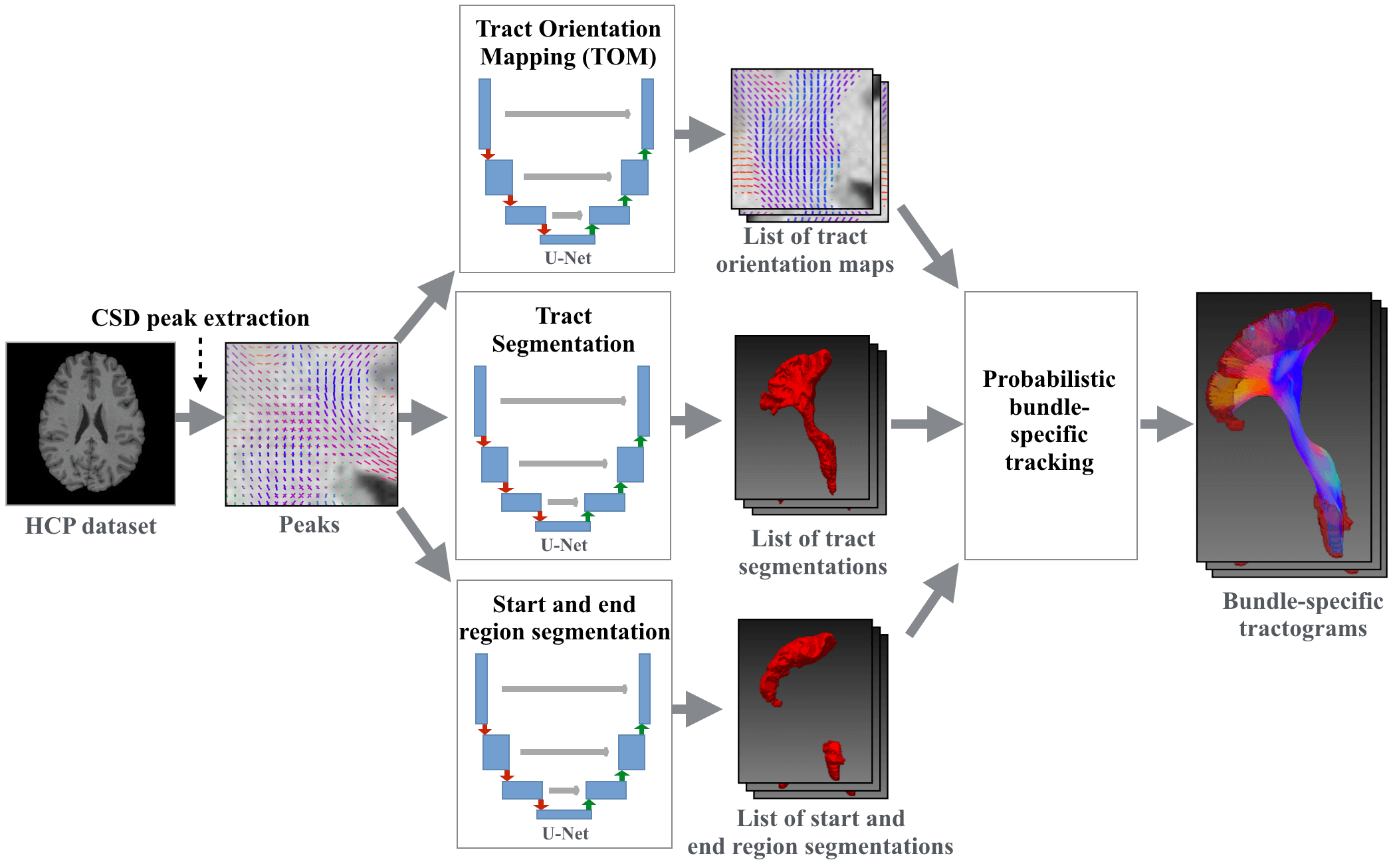}
		\caption{Pipeline overview: Constrained spherical deconvolution (CSD) is applied to obtain the three principal FOD directions per voxel which is the input to three U-Nets. The three U-Nets are used to create a tract orientation map, a tract mask and a start/end region mask for each tract. Then probabilistic tracking is run on the tract orientation maps. All streamlines leaving the tract mask and not ending in the start/end masks are discarded. The result is one tractogram for each tract.}
		\label{fig:overview}
	\end{figure*}

	In \citet{Wasserthal18b} we presented a novel concept called tract orientation mapping (TOM) that approaches the problem before doing tractography by learning tract-specific peak images (tract orientation maps, also abbreviated TOM). Each TOM represents one tract, and each voxel contains one orientation vector representing the local tract orientation, i.e. the local mean streamline orientation of the tract (see Fig. \ref{fig:peaks}). These tract orientation maps can then be used as a prior -- similar to \citet{rheault_bundle-specific_2018}, who employed registered atlas information as a tract-specific prior -- or directly as orientation field for tractography. In \citet{Wasserthal18a} we presented a novel method called TractSeg for fast and accurate tract segmentation.
	Based on these preliminary works we introduce an comprehensive approach to bundle-specific tractography: 
	
	On low resolution data, TOM tends to oversegment the individual tracts. In contrast to the complex task of voxel-wise peak regression with TOM, the simpler binary segmentation with TractSeg yields more accurate tract delineations. Therefore we use the segmentation results from TractSeg to filter the TOM tractograms.
	After filtering with the TractSeg segmentations the tractograms show good spatial extent and orientation. However, a lot of streamlines are still ending prematurely. Filtering the streamlines by a gray matter segmentation is not sufficient, as tracts tend to touch gray matter regions but are not supposed to end there. To obtain proper segmentations of the regions where each tract starts and ends, we trained another convolutional neural network using the same approach as TractSeg \citep{Wasserthal18a}. Now all streamlines not ending in the start/end regions can easily be removed.
	Using the steps described so far, highly accurate bundle-specific tractograms can be obtained in most situations. However, in some cases a simple deterministic tracking of the TOM peaks yields sub-optimal results, for example due to low image resolution. Therefore we propose a probabilistic approach to TOM tractography which maximizes the sensitivity of the proposed bundle specific tractography pipeline, even on low resolution data or strongly bent tracts. An overview of the entire pipeline can be seen in Fig. \ref{fig:overview}.
	
	For training and the first part of the evaluation we use the dataset provided by \citet{Wasserthal18a} containing 105 subjects from the Human Connectome Project (HCP). For the second part of our evaluation we use 17 differently acquired datasets to evaluate how good our approach generalizes to other datasets. We compare our method to seven other state-of-the-art methods for generating bundle-specific tractograms. We show that our approach is easy to set up, fast to run and does not require affine or elastic registration, parcellation or clustering.
	
    In comparison to our previous works \citep{Wasserthal18a, Wasserthal18b} this paper adds the following contributions: Segmentation of the tract start/end region, combination of tract segmentation, start/end region segmentation and tract orientation maps, more sensitive custom tracking algorithm which is optimized for this approach and extended evaluation on more bundles, phantom data and 17 non-HCP datasets with and without pathologies.
    
	A short note on the terminology we use: When talking about \textit{fibers} or \textit{streamlines} we are referring to the single streamlines. When talking about \textit{bundles} or \textit{tracts} we are referring to a group of streamlines making up an anatomical structure (e.g. the corticospinal tract).

	\section{Materials and Methods}
	All three methods (tract segmentation, start/end region segmentation and tract orientation mapping) are based on the same fully convolutional neural network architecture (U-Net \citep{ronneberger_u-net_2015}) that receives as input the fiber orientation distribution function (FOD) peaks. What differs is the training target the network has to learn. For tract segmentation, the network is performing voxel-wise binary classification to discern tract and non-tract voxels. For tract start/end region segmentation it is also doing binary classification but now the number of classes has doubled because for each tract one start and one end region is learned. For tract orientation mapping the network regresses a single 3D peak vector, i.e. three float values, per voxel and bundle. In this way, the models used for the three methods only differ in the number of output channels, the final activation function and the loss function.
	
	\subsection{Preprocessing}
	While we successfully tested raw diffusion weighted images as input for our method, this would have restricted the method to the MRI acquisition used during training, not allowing for any variation in the acquisition, such as a change of b-value or the number of gradient directions, without a complete retraining of the model. Moreover, for high angular resolution datasets, it would have resulted in an input image with an accordingly large number of channels (one channel for each gradient orientation and each b-value), resulting in unfeasible high memory demand and slow file input/output during training. A more condensed representation of the data was chosen to mitigate this problem: The network expects to receive the three principal fiber directions per voxel as input, thus requiring nine different input channels (three per principal direction). In this study, the principal directions were estimated from the diffusion data using the multi-shell multi-tissue constrained spherical deconvolution (CSD) and peak extraction available in MRtrix \citep{jeurissen_multi-tissue_2014, tournier_robust_2007} with a maximum number of three peaks per voxel. If a voxel contained only one fiber direction e.g. voxels in the corpus callosum, then the second and third peak are set to zero. Another possible input instead of peaks would have been FA maps. However, this resulted in worse results (see supplementary materials).
    The HCP images have a spatial resolution of $145 \times 174 \times 145$ voxels. We cropped them to $144 \times 144 \times 144$ without removing any brain tissue to make them fit to our network input size.
	
	\subsection{Model}
	\subsubsection{Architecture}
	The proposed 2D encoder-decoder architecture was inspired by the U-Net architecture previously proposed by \citet{ronneberger_u-net_2015}. To enable better flow of the error gradients during backpropagation we added deep supervision \citep{Isensee18}. This reduced the training time and slightly improved the results. A figure of the network architecture can be found in the supplementary materials.
	
	The input for the proposed network is a 2D image at $144 \times 144$ voxels and 9 channels corresponding to the 3 peaks per voxel (each 3D peak is represented by three float values). The output is a multi-channel image with spatial dimensions of $144 \times 144$ voxels, where each channel contains the voxel-wise results for one tract.  For tract segmentation this leads to 72 channels and for start/end region segmentation to $72\cdot2=144$ channels. Following the same approach $72\cdot3=216$ channels would have been needed for tract orientation mapping. However, given such a high number of classes the training did not converge anymore. To deal with this issue we chose to only train for 18 tracts (=54 channels) at the same time. So four models had to be trained to cover all 72 tracts. \\
	For the segmentation tasks the networks output a probability between 0 and 1. These probabilities are converted to binary segmentations by thresholding at 0.5.  For the tract orientation mapping the networks output one peak per voxel and tract. Peaks shorter than 0.3 are discarded. \\
	To avoid a downsized output in comparison to the input we padded with half the filter size (rounded down). Given a filter size of 3 the padding was set to 1. This is also referred to as SAME padding \citep{dumoulin_guide_2016}.
	
	\subsubsection{Handling of 3D data}
	While in principle the U-Net architecture allows extensions to image segmentation with 3D convolutions \citep{cicek_3d_2016}, we here propose a 2D architecture. Using a 3D U-Net, we did not achieve the same performance as when using a 2D U-Net (see supplementary materials). To still leverage the additional information provided by the third dimension, we randomly sampled 2D slices in three different orientations during training: axial, coronal and sagittal. This meant that our model learned to work with all three of these orientations. During inference three predictions per voxel per tract were generated, one for each orientation, resulting in an image with dimensions of $144\times144\times144\times nr\_classes \times 3$ (after running our model $144\cdot3$ times). We use the mean to merge those three prediction to one final prediction. Running the model three times (once for each orientation) is only done for tract segmentation and start/end region segmentation. For tract orientation mapping it slightly worsened the results (+0.18 angular error degrees on the HCP Quality dataset). Therefore we only run the model once for all slices along the y-axis. The y-axis corresponds to the coronal axis and as shown in \citet{Wasserthal18a}, the y-axis gives the best results when only using one axis.

	\subsection{Training}
	\label{sec:training}
	
	\subsubsection{Loss}
	For the segmentation models we trained our network using the binary cross-entropy loss. Sigmoid activation functions were used in the last layer. For a given target $y$, an output of the network $\hat{y}$ and $N$ number of classes, the loss is calculated as follows:
	
	\begin{equation}
		loss(\hat{y}, y) = -\frac{1}{N} \sum_{i=0}^{N} (y[i]log(\hat{y}[i]) + (1 - y[i]) log(1- \hat{y}[i])]
	\end{equation}
	
	For the tract orientation mapping model the network was trained using cosine similarity as loss. Linear activation functions were used in the last layer. The loss is defined as follows
	
	\begin{equation}
		loss(\hat{y}, y) = - \frac{1}{N} \sum_{i=0}^{N}  \frac{\abs{\langle\hat{y_{i}}, y_{i}\rangle}}{\norm{\hat{y_{i}}}_{2} * \norm{y_{i}}_{2}}
	\end{equation}
	with $N$ being the number of classes, $y$ the training target and $\hat{y}$ the network output. In \citet{Wasserthal18b} we used the cosine similarity in combination with the peak length as loss, thus allowing the model to also learn the extent of each bundle. However, learning the tract segmentation and the peak angles in two separate models is giving better results (for details see supplementary materials). Therefore we only use the cosine similarity in the loss of the tract orientation mapping model.
	
	\subsubsection{Hyperparameters}
	Leaky rectified linear units (ReLU) were used as nonlinearity \citep{nair_rectified_2010}. A learning rate of 0.001 was used and Adamax \citep{kingma_adam:_2014} was chosen as an optimizer. When the validation loss did not decrease for at least 20 epochs the learning rate was reduced by one order of magnitude. The batch size was 47. All hyperparameters were optimized on a validation dataset independent of the final test dataset. The network weights of the epoch with the highest Dice score during validation were used for testing.
	
	\subsubsection{Data augmentation}
	\label{sec:data_augmentation}
	To improve the generalizability of our model, we applied heavy data augmentation to the peak images during training \footnote{https://github.com/MIC-DKFZ/batchgenerators}. The following transformations were applied to each training sample. The intensity of each transformation was varied randomly by sampling from a uniform distribution $U$.
	
	\begin{itemize}
		\item Rotation by angle $\varphi_x \sim  U[-\pi / 4, \pi / 4]$, \newline $\varphi_y \sim  U[-\pi / 4, \pi / 4]$, $\varphi_z \sim  U[-\pi / 4, \pi / 4]$
		\item Elastic deformation with alpha and sigma \newline $(\alpha, \sigma) \sim (U[90,120], U[9,11])$. A displacement vector is sampled for each voxel $d \sim U[-1, 1]$, which is then smoothed by a Gaussian filter with standard deviation $\sigma$ and finally scaled by $\alpha$.
		\item Displacement by \newline $(\Delta x, \Delta y) \sim (U[-10,10], U[-10,10])$
		\item Zooming by a factor $\lambda \sim  U[0.9, 1.5]$
		\item Resampling (to simulate lower image resolution) with factor $\lambda \sim  U[0.5, 1]$
		\item Gaussian noise with mean and variance \newline $(\mu, \sigma) \sim (0, U[0, 0.05])$
	\end{itemize}
	
	The training samples were normalized to zero mean and unit variance before passing them to the network.
	When training our network on peaks generated by the MRtrix multi-shell multi-tissue CSD method, we found that it did not work well on peaks generated by the standard MRtrix CSD method. In order to ensure our model worked well with all types of MRtrix peaks, we generated three peak images: (1) multi-shell multi-tissue CSD using all gradient directions, (2) standard CSD using only $b=1000s/mm^2$ gradient directions, (3) standard CSD using only 12 gradient directions at $b=1000s/mm^2$.
	During training, we randomly sampled from these three peak images, thus ensuring that our network worked well with all of them.
	We trained for 250 epochs with each epoch corresponding to 193 batches. This means that over the course of the entire training, the network has seen 2,267,750 slices which have been randomly sampled from axial, coronal and sagittal orientations, randomly sampled from three different peak types and randomly permutated by the data augmentation transformations.
	The results presented in section \ref{sec:experiments} were obtained using an implementation of the proposed method in Pytorch \footnote{pytorch.org}.
	
	\subsubsection{Super resolution}
	Our models were trained with images of size $144 \times 144$ corresponding to the 1.25mm resolution of the HCP data. As mentioned in section \ref{sec:data_augmentation} we were using resampling as data augmentation. This means images were downsampled to a resolution of 2.5mm to simulate lower resolution images. Then they were upsampled back to 1.25mm to fit the $144 \times 144$ input size of the model. So the resolution kept the same but the images got blurred by the down- and upsampling. This down/upsampling was only done for the input images (peaks) not for the labels (training target). This way the models were able to learn a higher resolution output than was actually provided as input. This is commonly referred to as \textit{super resolution} \citep{alexander_image_2017}.
	When our approach receives a low resolution image as input, it is first upsampled to resolution 1.25mm and then fed to the model which returns a output also in 1.25mm resolution. This higher resolution especially helps on very thin bundles like the anterior commissure (CA).

	\subsection{Data}
	For training our models the dataset published by \citet{Wasserthal18a} was used. It contains reference delineations of 72 major white matter tracts (see supplementary materials for a list of all tracts) in 105 subjects from the Human Connectome Project. The details of how this dataset was curated are described in \citet{Wasserthal18a}.
	The reference delineations are provided in form of streamlines. In this paper we refer to this dataset as \textit{reference data} or \textit{reference tracts}.
	
	\subsubsection{Preprocessing of reference data for different tasks}
	\label{sec:data_processing}
	To be able to use the dataset for our three tasks (tract segmentation, start/end region segmentation and tract orientation mapping) some preprocessing was necessary: \\
	For the tract segmentation we convert the reference streamlines to binary masks by setting each voxel to $True$ where at least one streamline runs through. \\
	For the start/end region segmentation we create binary masks from the streamlines start and end points. However, streamlines have no defined direction. So for example for the corticospinal tract some streamlines start at the cortex whereas other streamlines start at the brain stem. Therefore the start point of one streamline might be in the same region as the end point of the next streamline. The resulting binary mask is the union of the start and end region of a tract. Splitting the union into two binary masks, one for the start and one for the end region is not trivial as for some tracts like the uncinate fasciculus those region can be very close together. To avoid manual separation (which is time consuming and less objective) we took the following approach for splitting the regions: First we used a clustering algorithm (DBSCAN \citep{ester_density-based_1996}) to create two clusters from the combined region. The clustering was only done on a subset of the data points to avoid long runtimes. When the start and end region were close together the clustering sometimes misassigned points. Therefore we used the results from the clustering to train a random forest. This led to a correct separation of the two region for all subjects and ensured fast runtime when running for all data points. From the points in those two regions binary masks were created. Finally we did binary closing and a small amount of binary dilation using scipy \citep{jones_scipy_2001} to create a consistent region from the single points. \\
	For the tract orientation mapping the main streamline orientation in each voxel had to be determined for each tract. Using the mean of all streamlines running through a voxel led to rather noisy results. Therefore we used Mean Shift clustering to group the orientations of all streamlines in one voxel. Then the mean of the orientations in the biggest cluster was taken as final orientation for that voxel. This substantially reduced the noise.
	
	\subsubsection{Clinical quality dataset}
	The reference dataset is provided in high HCP data quality  (\textit{HCP Quality}). However, in clinical routine, faster MRI protocols are used which result in lower quality data. To test how the proposed method performs on clinical quality data, we downsampled the HCP data to 2.5 mm isotropic resolution and removed all but 32 weighted volumes at $b=1000 s/mm^2$. We call this dataset \textit{Clinical Quality}. The reference tracts from the \textit{HCP Quality} dataset were reused as our reference tracts here. This provides high quality reference tracts for the low quality data, thus allowing proper evaluation. 
	
	\subsubsection{Phantom dataset}
	The \textit{Clinical Quality} dataset has lower resolution and less directions than the \textit{HCP Quality} dataset but it was still acquired by the same scanner. To evaluate how the proposed method generalizes to images from other scanners and other acquisition settings we would need a dataset with reference tract delineations from another scanner. Unfortunately such a dataset is not available and using the same approach as was used for the \citet{Wasserthal18a} dataset is not feasible: For lower quality datasets it becomes very difficult and ambiguous for an expert to accurately determine where tracts run. The expert delineations would rather be approximations not suitable for detailed evaluation. One solution, however, is to simulate low quality data from a different scanner. Thereby we have perfect ground truth and still low image quality. We used the toolkit FiberFox \citep{neher_fiberfox:_2014} to create such software phantoms.
	We selected 21 subjects (not used for training) from the reference data and for each simulated the diffusion weighted image of a brain containing only the 72 reference tracts. 
	The simulated images have an isotropic resolution of 2.5mm, 32 gradient directions at $b=1000mm/s$ and several artefacts which were randomly chosen from the following list: head motion, ghosts, spikes, eddy currents, ringing, distortions, signal drift and complex Gaussian noise. We call this dataset \textit{Phantom}.
	As the \textit{Clinical quality} dataset and the \textit{Phantom} dataset only have one b-value shell, we cannot use multi-shell CSD as we did for the \textit{HCP Quality} data. Instead MRtrix standard CSD was used to generate the peaks of the FOD.

	\subsection{Bundle-specific tractography}
	
	\subsubsection{Flavors of TOM}
\begin{figure*}[!t]
		\centering
		\includegraphics[width=13cm]{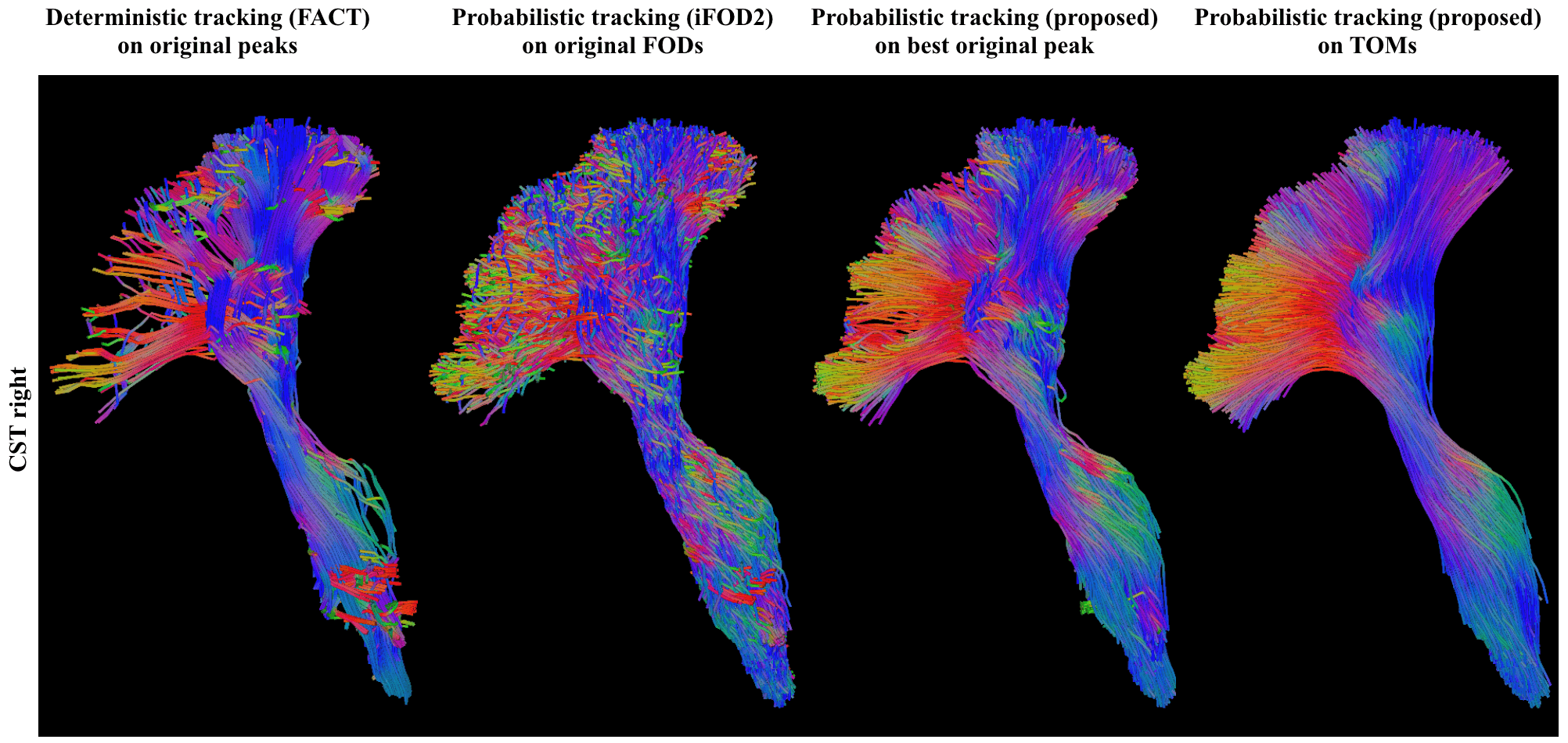}
		\caption{Right corticospinal tract (CST) in one subject from the BrainGluSchi \citep{bustillo_glutamatergic_2017} dataset (2mm isotropic resolution, 30x $b=800mm/s^2$) reconstructed by the different tracking variants. Probabilistic tracking on TOMs shows the best results in terms of sensitivity and specificity.}
		\label{fig:tracking_algos}
	\end{figure*}

	There are three different ways how tract orientation maps can be used to create bundle-specific tractograms:
	\begin{itemize}
		\item Directly track on the tract orientation maps
		\item For each voxel select the peak from the original input peaks which is closest to the orientation predicted by TOM. Then track on these peaks. This has the advantage of staying closest to the original signal, but if the original peaks are quite noisy the chosen peak will also be noisy. This is a problem especially on low quality data.
		\item Use the tract orientation map as a prior by taking the weighted mean between the predicted orientation from the TOM and the original orientation normally used for tracking.
	\end{itemize}
	Fig. \ref{fig:tracking_algos} shows exemplary results for the different tracking options on one subject from a low resolution dataset. In all four cases the tract masks as well as the start/end region masks were used to filter the tractograms. Tracking on the original signal is insufficient: Deterministic tracking lacks sensitivity whereas probabilistic trackings lacks specificity (many false positives). Tracking on the tract orientation maps gives the best of both: high sensitivity (tract is complete) and high specificity (few false positives). Tracking on the best original peaks also shows good results but is missing small parts of the lateral projections of the CST.
	Therefore for our experiments we chose the first option: Directly track on the tract orientation maps. This gave the best results, especially on low quality data where the original peaks can be quite noisy.
	
	\subsubsection{Probabilistic tracking on peaks} 
\begin{figure*}[!t]
		\centering
		\includegraphics[width=13cm]{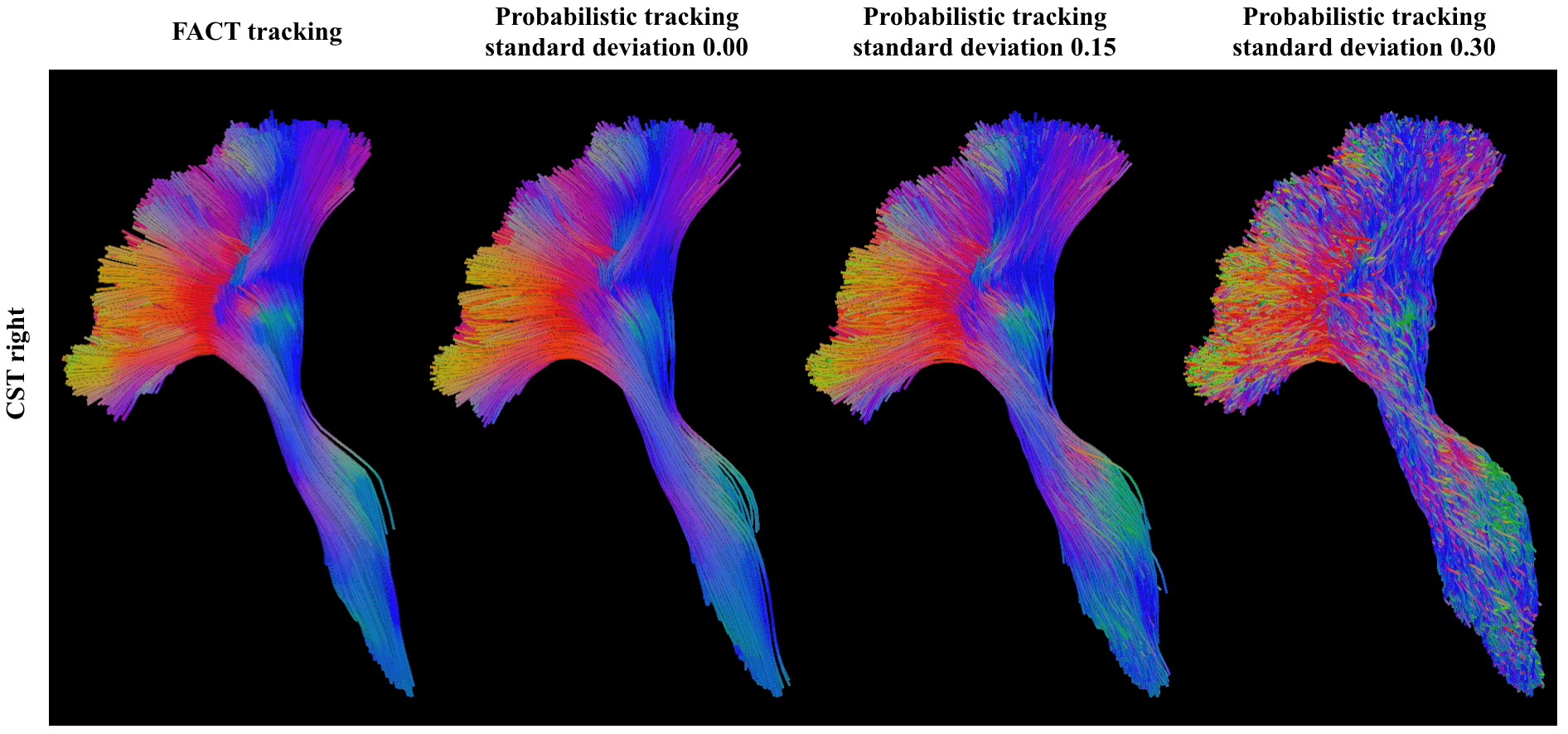}
		\caption{Reconstructions of the right corticospinal tract (CST) in one subject from the BrainGluSchi \citep{bustillo_glutamatergic_2017} dataset (same subject as in Fig. \ref{fig:tracking_algos}. Comparison of deterministic FACT tracking on tract orientation maps (TOMs) with proposed probabilistic tracking with different standard deviations for the Gaussian distribution used for sampling. Up to a standard deviation of 0.15 the probabilistic tracking results are very consistent with the FACT tracking. Using higher values leads to clearly more spurious streamlines.}
		\label{fig:comparison_prob_stddev}
	\end{figure*}

	The output of the tract orientation mapping is one tract orientation map for each tract. A tract orientation map contains one 3D vector (one peak) at each voxel telling the main orientation of the respective tract at that voxel. Creating streamline from these maps can easily be done by using deterministic tractography (e.g. \citet{mori_three-dimensional_1999, basser_vivo_2000}). This works well on high resolution data. However, on low resolution data just following the main orientation in each voxel sometimes leads to small branchings being missed as they cannot be represented on the low resolution. Probabilistic tractography enables more sensitive tracking. By not just following the main orientation in each voxel but sampling from the orientation distribution, smaller branchings can be reconstructed that otherwise would be missed. In our case, however, only one orientation per voxel is provided by the tract orientation map, but no orientation distribution. To be able to sample from orientations around the main orientation, we use a Gaussian distribution centered on the main orientation with a fixed standard deviation. 
    When using a small value for the standard deviation this can be interpreted as modeling a lower bound for the FOD dispersion, as in real data the width of the FOD lobes of all major tracts is non-zero. Even in areas of highly consistent fiber orientation like the corpus callosum there still is dispersion. For the fixed standard deviation we chose a value (0.15) which leads to less dispersion than in the corpus callosum FODs (see supplementary materials for more details). This way the fixed standard deviation is a conservative estimate of the lower bound of the dispersion and helps to increase sensitivity compared to using plain deterministic tracking.\\
    Trying to learn a complete orientation distribution for each voxel instead of only learning the main orientation and using a global lower bound for the dispersion could be promising. However, experiments in this direction did not lead to better tracking results. Therefore we decided to stick with the easier approach of an empirically determined fixed standard deviation.\\
    Although our approach is probabilistic it is quite different from the algorithms commonly referred to as probabilistic tractography e.g. \citet{behrens_characterization_2003}. For those probabilistic tracking algorithms the orientation distribution is representing the uncertainty in the underlying signal. In our approach the probabilistic component is only a way for increasing sensitivity to capture small branchings (relating it to \citet{ankele_versatile_2017}) by assuming a lower bound on the fiber orientation uncertainty.\\
	Our tracking algorithm is based on the deterministic algorithm described by \citet{basser_vivo_2000} with the main difference that at each step the next orientation to take is sampled from the given Gaussian distribution. Then the tracking algorithm is taking a step (with fixed step size) along this sampled orientation. At the end the streamlines are interpolated using b-splines.
	All streamlines have to start and end in the regions segmented by the start/end region segmentation model and are not allowed to leave the mask generated by the tract segmentation model, otherwise they are discarded. \\
    Using probabilistic tractography increases sensitivity but this often comes at the cost of an increased number of false positives. In our case we can keep the number of false positives introduced by the probabilistic tracking quite small as the tracking is highly constrained by the tract mask as well as the start/end region mask. Moreover we use a Gaussian distribution with a quite small standard deviation of 0.15. As can be seen in Fig. \ref{fig:comparison_prob_stddev} using a higher value like 0.3 would lead to many spurious fibers. But using a value of 0.15 results in trackings which are highly consistent with deterministic streamline tractography, showing that our approach is close to the specificity of deterministic tractography while being more sensitive. \\
	During tracking the following parameters were used: a step size of 0.7 voxels and a minimum streamline length of 50mm. Seeds were randomly placed inside of the tract mask until a maximum of 2000 streamlines per tract were created.
	
	\subsection{Reference methods}
	We compared our proposed method to 3 methods for automatic tract delineation (comparing segmentation performance in terms of DICE score and orientation quality in terms of voxel-wise angular error): TractQuerier \citep{wassermann_white_2016}, RecoBundles \citep{garyfallidis_recognition_2017} and streamline atlas. 
	Moreover we compared to 3 methods for tract segmentation (comparing only segmentation performance): custom atlas registration, FSL atlas registration and multiple mask registration. 
	We also compared to 2 methods which give a voxel-wise orientation for each tract (comparing only orientation quality): Peak atlas and the best original peak. 
	These methods include clustering-based as well as ROI-based approaches. We give an outline of how they work (1.) and how we applied them (2.).
	
	\subsubsection{TractQuerier}
	\label{sec:TractQuerier}
	1. TractQuerier \citep{wassermann_white_2016} extracts tracts based on the regions the streamlines have to start at, end at and (not) run through. 2. We compared our method to the output from TractQuerier using the same queries as used in \citet{Wasserthal18a} without any further post-processing. The queries were applied to a whole brain tracking generated with MRtrix (Tournier et al., 2010) and the following settings:
	Constrained spherical deconvolution (for \textit{HCP Quality} data using the multi-shell multi-tissue option) was used to extract the FOD and probabilistic tractography (iFOD2) (for \textit{HCP Quality} data using the anatomically constrained option) was used to generate a whole brain tractogram. For the \textit{HCP Quality} data 10 million streamlines were generated, for all other datasets 500,000 streamlines. The minimum length was set to 40 mm. The other parameters were kept at their default values.
	As parcellation the freesurfer Deskian/Killiany atlas was used \citep{desikan_automated_2006}. Freesurfer was applied using the default settings.
	\subsubsection{RecoBundles}
	1. Given streamlines of a reference tract in a reference subject, RecoBundles \citep{garyfallidis_recognition_2017} can be used to find the corresponding streamlines in a new subject. 2. We randomly picked 5 reference subjects from the training dataset. Due to the long runtime for RecoBundles, a higher number of reference subjects was not feasible. Then we ran RecoBundles 5 times for the new subject (once for each reference subject) using the default RecoBundles parameters (see supplementary materials) and the same whole brain tractogram used for TractQuerier (see section \ref{sec:TractQuerier}). This resulted in 5 extractions of each tract in the new subject. To get a final segmentation, we took the mean of those 5 extractions.
	
	\subsubsection{Streamline atlas registration (SLAtlas)}
	1. Given streamlines of a reference tract in a reference subject, registration can be used to align them with a new subject. 2. The same 5 reference subjects as those selected for RecoBundles were used. To delineate the tracts in a new subject, we registered each of the 5 reference subjects to the new subject. Affine registration of the whole brain tractograms was already done by RecoBundles so we reused these transformations. Finally for each tract we merged the streamlines from the 5 registered reference subjects.
	
	\subsubsection{Atlas registration 1 (Atlas Custom)}
	1. Several subjects can be averaged to an atlas which can then be registered to new subjects to segment structures. 2. We split our dataset into training and testing data, using the same 5-fold cross-validation as used for the evaluation of our proposed method (see section \ref{sec:experiments}). The training data was used to create a tract atlas. Firstly, we registered all subjects to a random subject using symmetric diffeomorphic registration implemented in DIPY \citep{avants_symmetric_2008, garyfallidis_dipy_2014}. Registration was performed based on the FA maps of each image. After registration, the FA maps of all images were averaged. Then, in a second iteration all images were registered to this mean FA image. This two-stage approach limits the bias introduced by the initial subject choice in the first iteration. The tract atlas thus contained the tract masks for all 72 reference tracts. For each tract, we took the mean over all subjects, which produced a probability map. We thresholded the probability map at 0.5 to create a final binary atlas. During test time, the atlas was registered to the subjects of interest, yielding a binary mask for each tract in subject space.
	
	\subsubsection{Atlas registration 2 (Atlas FSL)}
	1. We compare to a second atlas method using a different implementation (FSL) and a different atlas template. 2. This method is identical to the previous method (Atlas Custom) except for the following two points: Instead of generating a mean FA template from the dataset we used the FA template provided by FSL (FMRIB58\_FA\_1mm). Instead of using the diffeomorphic registration of DIPY we used the linear (FLIRT) and nonlinear (FNIRT) registration of FSL \citep{jenkinson_fsl_2012}. As configuration we used the predefined configuration file for FA registration provided by FSL (FA\_2\_FMRIB58\_1mm.cnf).

	\subsubsection{Atlas registration 3 (Atlas MRtrix)}
	1. We compare to a third atlas method which is not based on registering FA images but on registering the FODs (fiber orientation distributions) thereby make use of the richer information of FODs compared to only using the FA. This method is implemented in MRtrix \citep{raffelt_symmetric_2011}. 2. This method is identical to the \textit{Atlas Custom} method except for the following two points: Instead of generating a mean FA template from the dataset we used the MRtrix method \textit{population\_template} to create a FOD template. Instead of using the diffeomorphic registration of DIPY we used the MRtrix nonlinear FOD registration. This method was not applicable for the \textit{Phantom dataset} as the FODs from the simulated Phantom data were not similar enough to the FODs from the FOD template to allow for meaningful registration.
	
	\subsubsection{Multiple mask registration (Multi-Mask)}
	1. Using an atlas can blur some of the details as it is based on group averages. The blurring can be reduced to some extent by registering the masks of single training subjects to a test subject instead of an averaged atlas. 2. The same 5 reference subjects as those selected for \textit{RecoBundles} were used. To segment the tracts in a new subject, we registered each of the 5 reference subjects to the new subject (symmetric diffeomorphic registration of the FA maps) and averaged the tract masks (from the reference tracts) of all 5 reference subjects. Finally, we thresholded this average at 0.5 to produce a binary mask for each tract in the space of the new subject. This differs from the \textit{Atlas registration} method in that the reference subjects are directly registered to subject space and are merged (1 registration) instead of first being registered to atlas space, then merging and being registered to subject space (2 registrations needed). Moreover, \textit{Atlas Registration} uses 63 subjects while \textit{Multi-Mask} only uses 5.
	
	\subsubsection{Peak atlas}
	This method is identical to \textit{Atlas FSL} with the only difference that instead of using binary masks we use peak images. The transformation calculated from registering the FA images is applied to each of the 3 peaks of the peak image independently using FSL vecreg which makes sure the peaks are reoriented accordingly \citep{alexander_spatial_2001}.
	
	\subsubsection{Best original peak (BestOrig)}
	1. Given the peak map of a reference tract, in each voxel we can choose the peak from the original signal that is closest to the peak from the reference tract, resulting in a new peak map. 2. For each subject in the test set we use the reference peaks to extract the best peak from the original signal. As we are using the ground truth in this method, it is not a fair method to directly compare to but it gives a good estimation of how good the original peaks are.

	\section{Experiments and results}
	\label{sec:experiments}
	
	\begin{figure*}[!t]
		\centering
		\includegraphics[height=6cm]{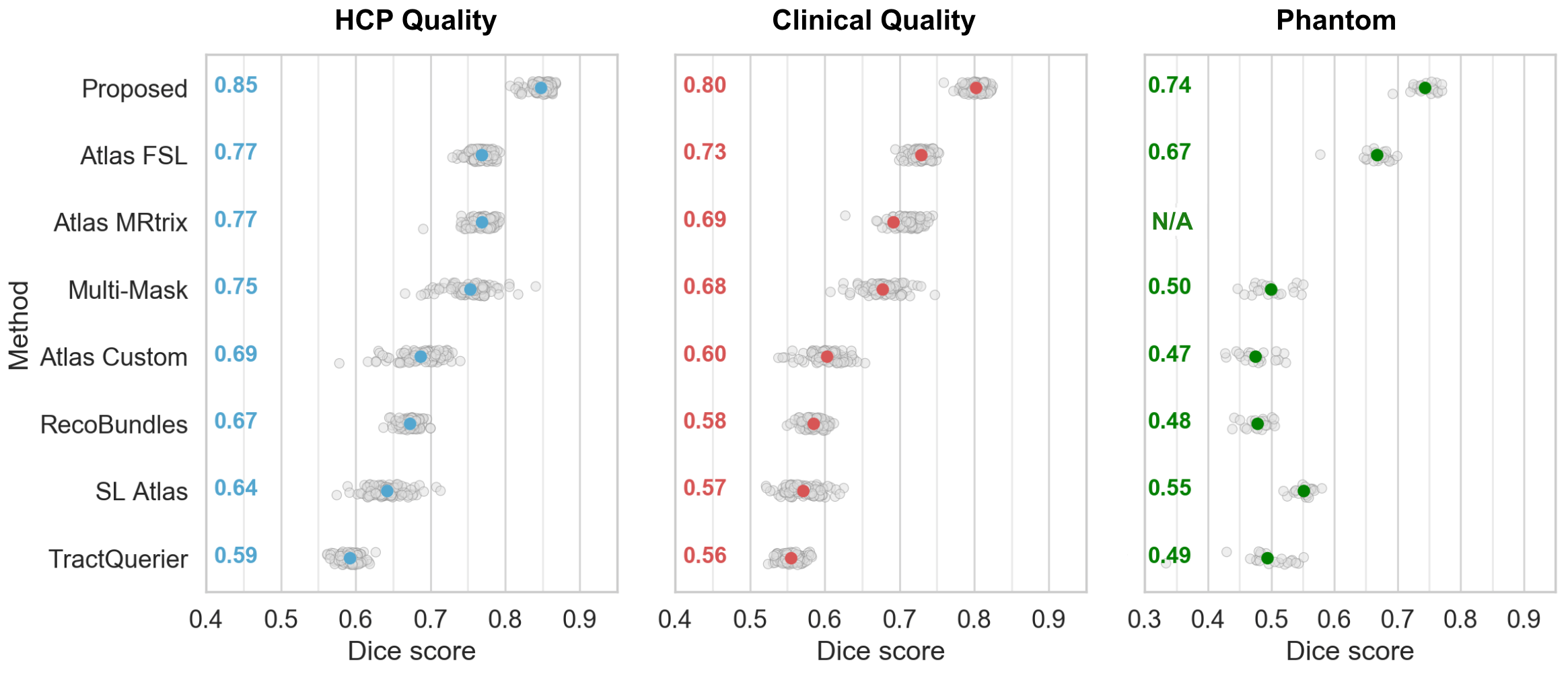}
		\caption{Segmentation results on the \textit{HCP Quality},\textit{ Clinical Quality} and \textit{Phantom} dataset with a gray dot per subject (mean over all tracts) and a colored dot for the mean over all subjects. Proposed: Our method; Multi-Mask: Multiple mask registration; Atlas: Atlas registration; SLAtlas: Streamline atlas.}
		\label{fig:segmentation_performance}
	\end{figure*}
	
	\begin{figure*}[!t]
		\centering
		\includegraphics[height=5cm]{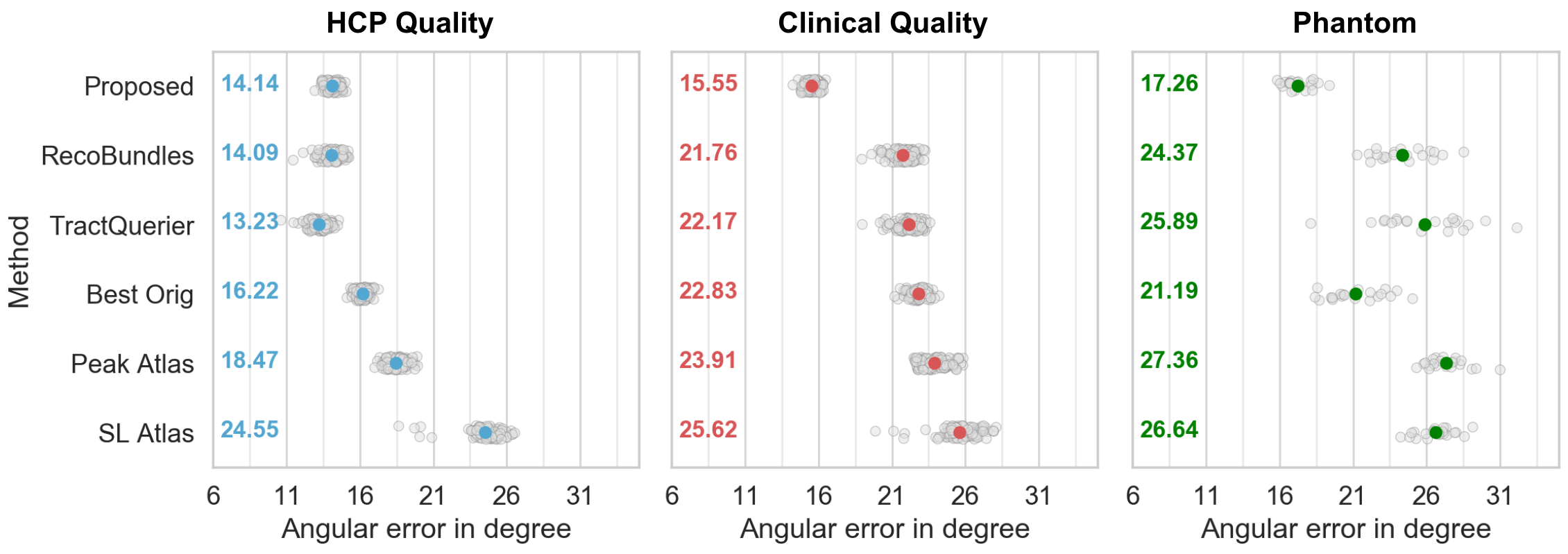}
		\caption{Orientation performance results on the \textit{HCP Quality},\textit{ Clinical Quality} and \textit{Phantom} dataset with a gray dot per subject (mean over all tracts) and a colored dot for the mean over all subjects. Proposed: Our method; BestOrig: Best original peak; SLAtlas: Streamline atlas.}
		\label{fig:orientation_performance}
	\end{figure*}
	
	For evaluation 5-fold cross-validation was used, i.e. 63 training subjects, 21 validation subjects (best epoch selection) and 21 test subjects per fold. The Wilcoxon signed-rank test \citep{wilcoxon_individual_1945} was used to test for statistical significance when comparing our method and the reference methods in the quantitative evaluation. For multiple testing, we applied the Bonferroni correction.

	\subsection{Segmentation performance}
	For evaluating segmentation performance we used the Dice score \cite{taha_metrics_2015} as our metric. The Dice score measures the overlap between two binary masks. It ranges from 0 to 1 with 1 being a perfect overlap. We calculated the Dice for each subject between each of the 72 reference tracts and the respective prediction of either our proposed method or one of the reference methods (e.g. RecoBundles). Then we averaged the Dice results for all 72 tracts to get one final Dice score per subject per method.
	Over all three datasets (\textit{HCP Quality}, \textit{Clinical Quality} and \textit{Phantom}) our proposed method significantly ($p < 0.01$) outperformed the reference methods by a large margin: on the \textit{HCP Quality} dataset it outperformed the reference methods on average by 14 Dice points and on the low quality datasets on average by 18 Dice points (\textit{Clinical Quality}) and 22 Dice points (\textit{Phantom}) (Fig. \ref{fig:segmentation_performance}).
	In general, the proposed method was less affected by the quality loss in the \textit{Clinical Quality} and \textit{Phantom} data than the reference methods.
	
	
	\subsection{Orientation performance}
	For evaluating orientation performance we use the voxel-wise angular error as metric. We calculate the angular error between the reference orientation and the orientation of the proposed method. We do this for every voxel where the reference peak and the peak of the proposed method have a length greater than zero. Then we average the errors to get one final angular error per subject per method.
	To calculate the voxel-wise main streamline orientation for the methods which output streamlines (\textit{RecoBundles}, \textit{TractQuerier} and \textit{SLAtlas}) we used the same technique as used for calculating the main streamline orientation for the reference data (see section \ref{sec:data_processing}): the streamline orientations in each voxel were first clustered and then the mean of the biggest cluster was chosen.
	On the \textit{HCP Quality} data \textit{RecoBundles} and \textit{TractQuerier} show slightly better orientation errors than our proposed method. Those methods have difficulties finding the borders of tracts (poor segmentation performance) but they are good at finding the correct streamlines belonging to the core of the tract and therefore show low orientation errors, as long as the underlying whole brain tractogram is of high quality. As soon as the image quality gets lower (\textit{Clinical Quality} and \textit{Phantom} dataset), the whole brain tracking also suffers and therefore the angular error of these methods rises significantly. Our proposed method on the other hand is not dependent on the whole brain tracking and quite robust to lower image quality as it was trained with extensive data augmentation. As a results the angular error only rises by 1 degree when using our proposed method on the \textit{Clinical Quality} data compared to the \textit{HCP Quality} data (Fig \ref{fig:orientation_performance}). 
	\textit{SLAtlas} and \textit{Peak Atlas} show high angular errors for all three datasets.
	
\begin{figure*}[!t]
		\centering
		\includegraphics[height=15cm]{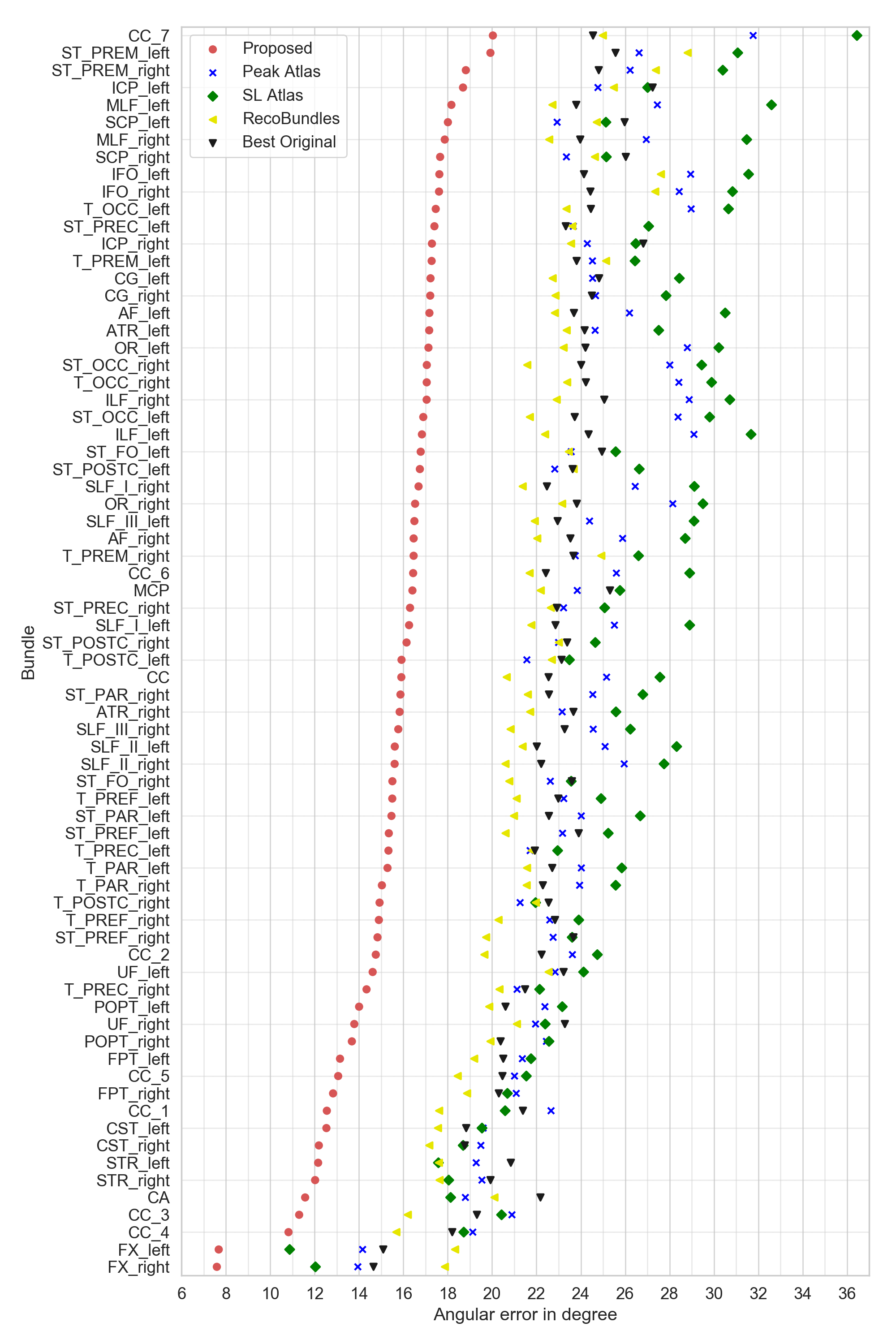}
		\caption{Angular errors for all 72 tracts on the \textit{Clinical Quality} dataset for our proposed method and all reference methods sorted by error. The following list shows the full names of each tract: 
		Arcuate fascicle (AF), Anterior thalamic radiation (ATR), Anterior commissure (CA), Corpus callosum (Rostrum (CC 1), Genu (CC 2), Rostral body (CC 3), Anterior midbody (CC 4), Posterior midbody (CC 5), Isthmus (CC 6), Splenium (CC 7)), Cingulum (CG), Corticospinal tract (CST), Middle longitudinal fascicle (MLF), Fronto-pontine tract (FPT), Fornix (FX), Inferior cerebellar peduncle (ICP), Inferior occipito-frontal fascicle (IFO), Inferior longitudinal fascicle (ILF), Middle cerebellar peduncle (MCP), Optic radiation (OR), Parieto-occipital pontine (POPT), Superior cerebellar peduncle (SCP), Superior longitudinal fascicle I (SLF I), Superior longitudinal fascicle II (SLF II), Superior longitudinal fascicle III (SLF III), Superior thalamic radiation (STR), Uncinate fascicle (UF), Thalamo-prefrontal (T\_PREF), Thalamo-premotor (T\_PREM), Thalamo-precentral (T\_PREC), Thalamo-postcentral (T\_POSTC), Thalamo-parietal (T\_PAR), Thalamo-occipital (T\_OCC), Striato-fronto-orbital (ST\_FO), Striato-prefrontal (ST\_PREF), Striato-premotor (ST\_PREM), Striato-precentral (ST\_PREC), Striato-postcentral (ST\_POSTC), Striato-parietal (ST\_PAR), Striato-occipital (ST\_OCC)
		}
		\label{fig:orientation_performance_per_bundle}
	\end{figure*}

	
	Fig. \ref{fig:orientation_performance_per_bundle} shows the angular error for each tract independently on the \textit{Clinical Quality} dataset.
	
	\subsection{Qualitative evaluation}
	For the qualitative evaluation, one subject (623844) was selected from the test set. We chose a subject whose Dice scores were closest to the mean Dice scores for the entire datasets to make the subject representative for the entire dataset. Since the scope of this manuscript does not allow us to show results for all 72 tracts, we selected three tracts that represent different degrees of reconstruction difficulty according to \citet{maier-hein_challenge_2017}: the inferior occipito-frontal fascicle  (IFO), corticospinal tract (CST) and anterior commissure (CA). The IFO is a tract which is fairly easy to reconstruct, which is reflected by its consistently good scores for all methods. The CST is more difficult to reconstruct. Its beginning at the brain stem is easy to reconstruct but as the fibers get closer to the cortex, they start to fan out. Finding these lateral projections is more difficult. Finally, the CA is a tract that is difficult to reconstruct. Due to its very thin body, it is hard to find streamlines running the entire way from the right to the left temporal lobe. The CA is one of the tracts with the lowest performance out of all of the methods.
	We show results for all reference methods that produce streamline output (\textit{RecoBundles}, \textit{TractQuier} and \textit{SLAtlas}). For each tract one 3D view is shown as well as one 2D slice allowing more in detail evaluation. On the 2D slice the mask of the bundle is shown (red) as well as the mask of the reference bundle (green).
	
	\begin{figure*}[!t]
		\centering
		\includegraphics[width=\textwidth]{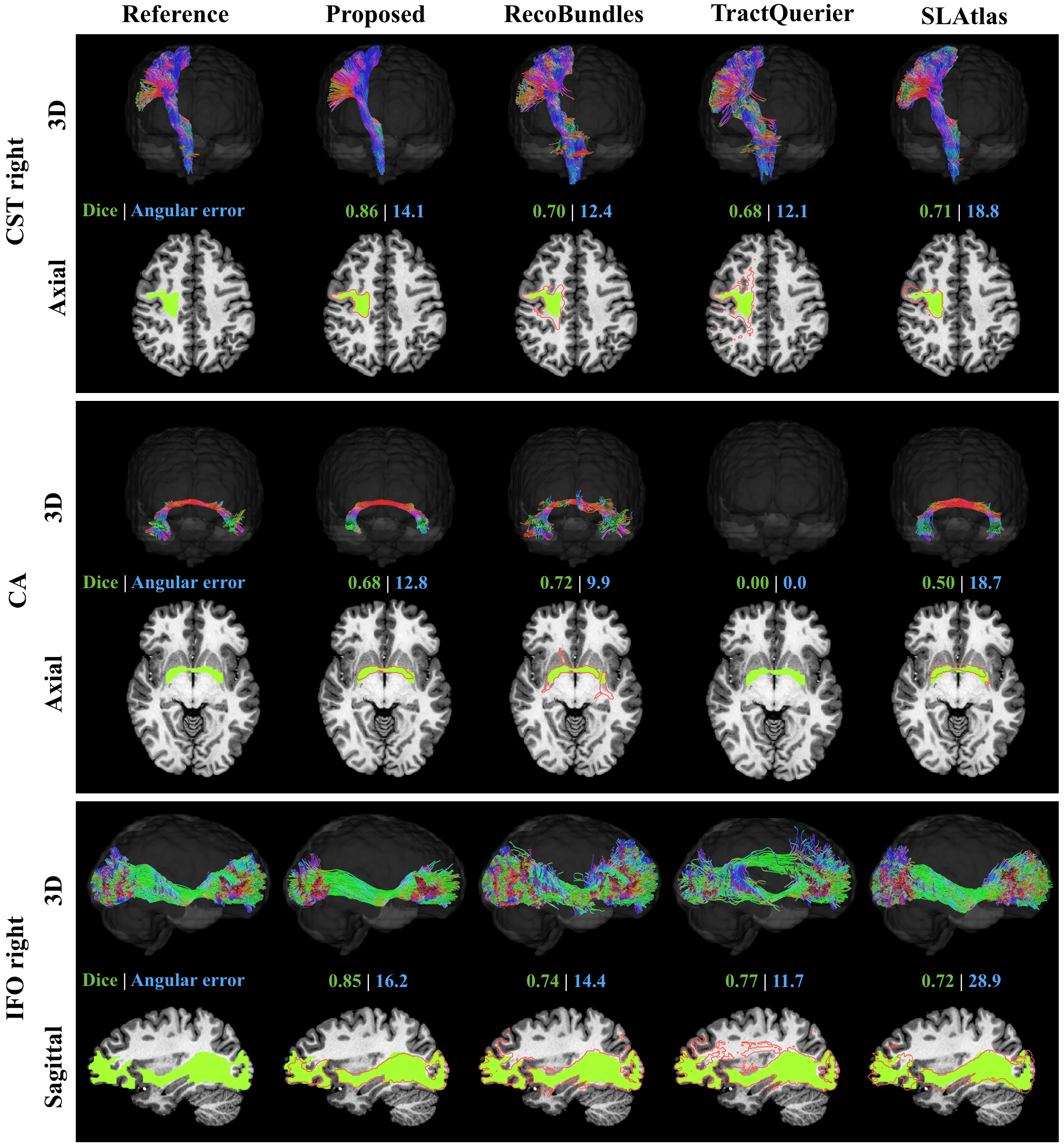}
		\caption{Qualitative comparison of results on \textit{HCP Quality} test set: reconstruction of right corticospinal tract (CST), anterior commissure (CA) and right inferior occipito-frontal fascicle (IFO) on subject 623844. Green shows the reference tract and red shows the tract mask of the respective method.}
		\label{fig:qualitative_results_hcp_quality}
	\end{figure*}
	
	As can be seen in Fig. \ref{fig:qualitative_results_hcp_quality}, the \textit{Proposed} method yielded accurate and spatially coherent reconstructions on all three tracts.  \textit{RecoBundles} oversegmented the CST to neighbouring gyri and selected many streamlines ending prematurely instead of reaching the correct start and end regions of the tract. \textit{TractQuerier} did not properly segment any of the example tracts. As it defines tracts mainly by their endpoints, it leaves much room for wrong turns between the start and end points. \textit{TractQuerier} extracts a lot of false positives, especially when using probabilistic tracking. The CA cannot be properly reconstructed with \textit{TractQuerier} as the default Freesurfer parcellation is not precise enough for the small parts of the CA. \textit{SLAtlas} produces reconstructions looking convincing on first sight but when looking at them on the 2D view we can see that it involves severe oversegmentation (e.g. segmenting gray matter and non-brain area for the CST) and slightly shifted tracts (e.g. CA). This is most probably owed to the affine registration, which cannot fully resolve the inter-subject variability.
	
	For the \textit{Phantom} dataset the different methods in principle show the same shortcomings as for the \textit{HCP Quality} dataset, but now more severely (see Fig. \ref{fig:qualitative_results_phantom} in the supplementary materials).

	\subsection{Generalization to other datasets and pathologies}
	To test the capability of the proposed method to generalize beyond HCP, which it was trained on, we applied it to 17 differently acquired datasets (including many public datasets like the OASIS, IXI, COBRE or Rockland datasets). A full list of all datasets can be found in the supplementary materials. These 17 datasets represent a wide variety of data: Different scanners, different spatial resolutions, different b-values, different number of gradients, healthy and diseased, normal and abnormal brain anatomy.\\
	
	\begin{figure*}[!t]
		\centering
		\includegraphics[width=\textwidth]{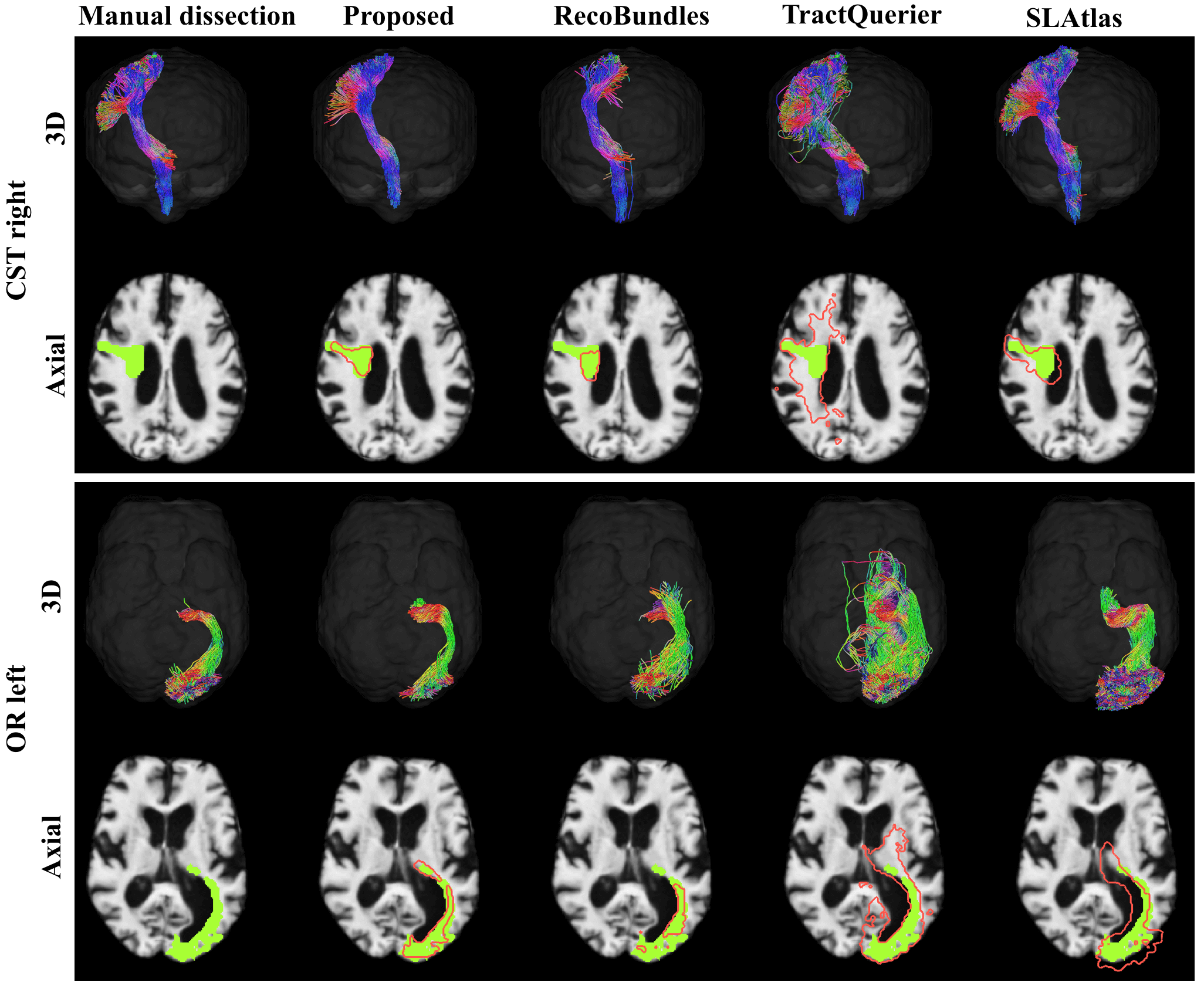}
		\caption{Qualitative comparison of results on one Alzheimer patient with enlarged ventricles from the OASIS dataset: reconstruction of right corticospinal tract (CST) and left optic radiation (OR). Green shows the manual dissection and red shows the tract mask of the respective method.}
		\label{fig:qualitative_results_big_ventricle}
	\end{figure*}
	
\begin{figure*}[!t]
		\centering
		\includegraphics[width=14cm]{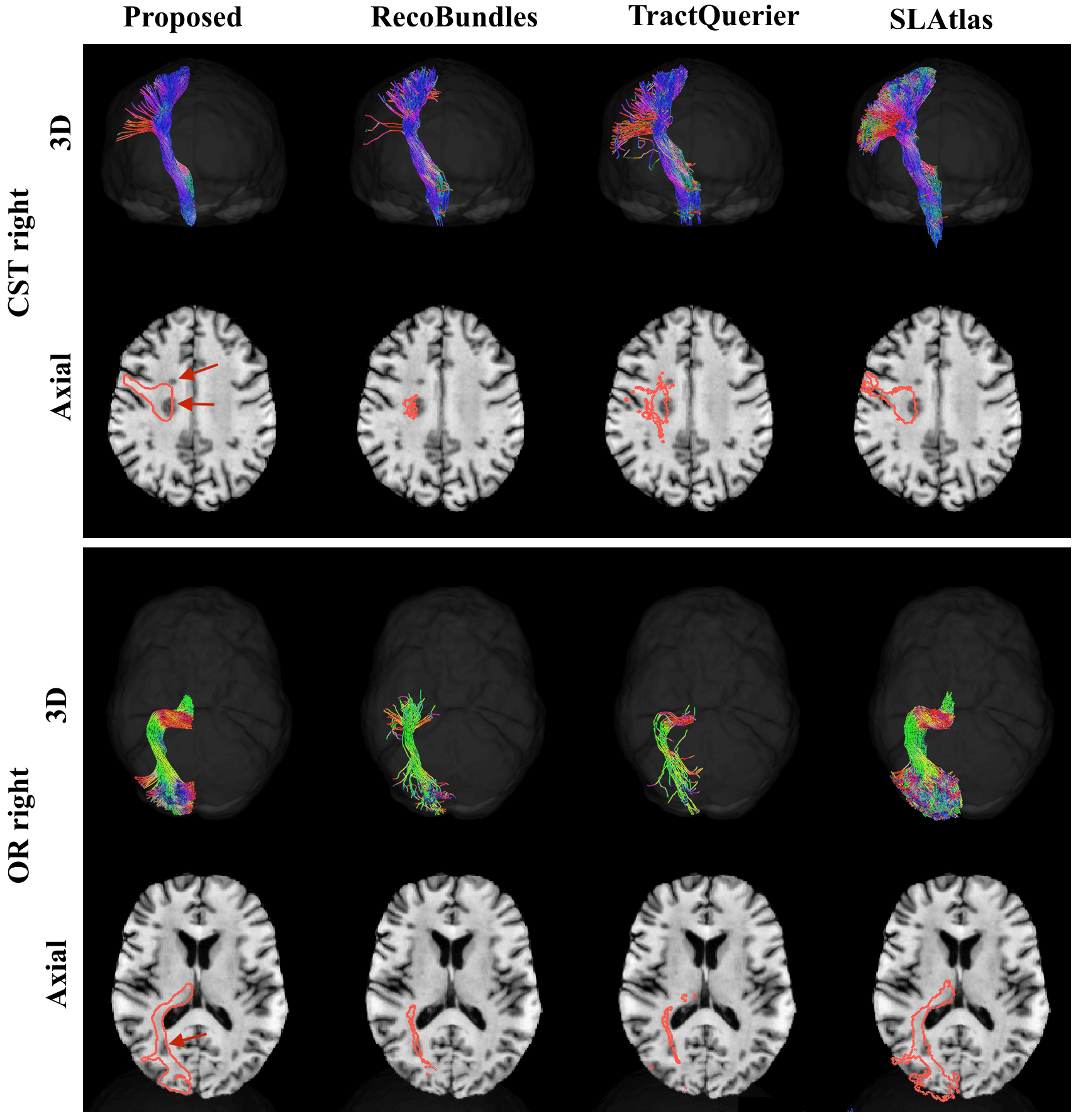}
		\caption{Qualitative comparison of results on one multiple sclerosis patient with several lesions inside of the tracts: reconstruction of right corticospinal tract (CST) and right optic radiation (OR). Red arrows show lesions close to the tracts.}
		\label{fig:qualitative_results_MS}
	\end{figure*}

\begin{figure*}[!t]
		\centering
		\includegraphics[width=14cm]{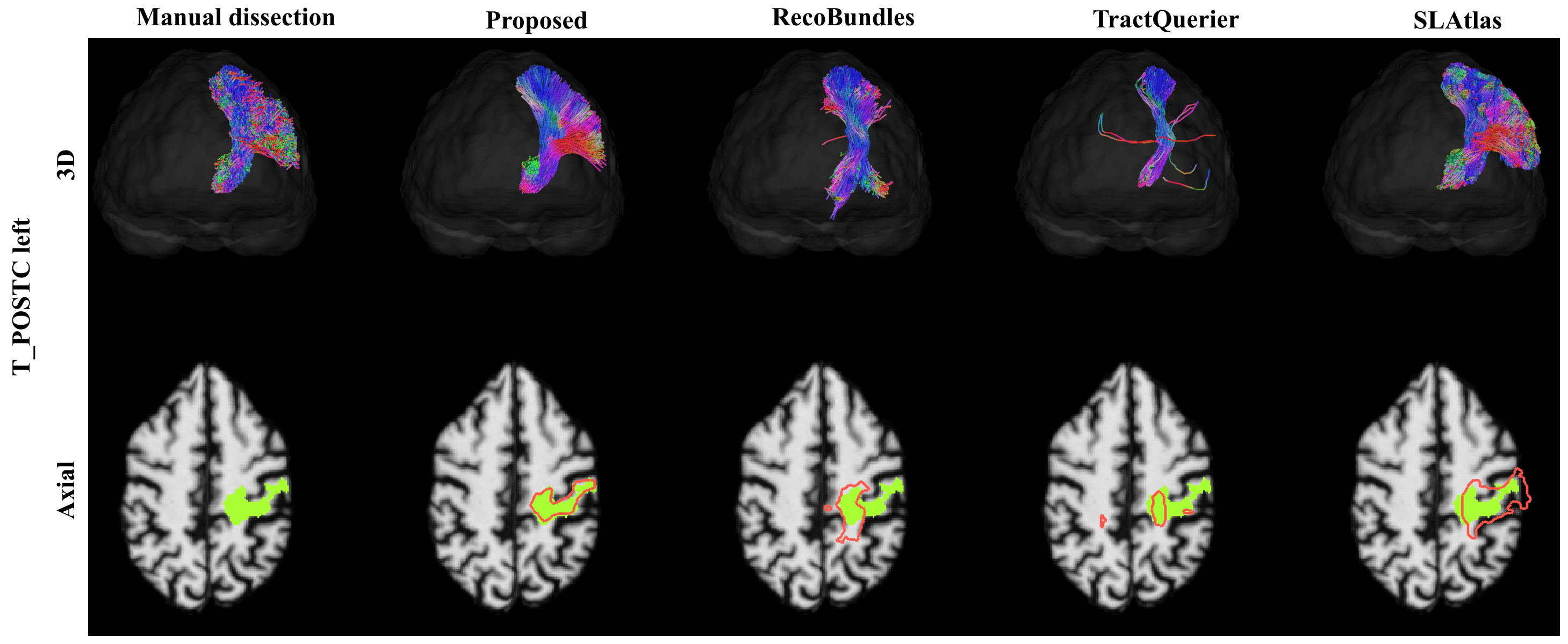}
		\caption{Qualitative comparison of results on one subject with brain volume loss and schizoaffective disorder: reconstruction of left Thalamo-postcentral tract (T\_POSTC). Green shows the manual dissection and red shows the tract mask of the respective method.}
		\label{fig:qualitative_results_atrophy}
	\end{figure*}

	An expert manually dissected the three tracts already shown in the above qualitative evaluation (CST, CA and IFO) from one randomly chosen subject from each dataset shown in table \ref{tab:datasets}. Visual comparisons were performed between manual dissections and the results of our proposed method as well as the previously introduced reference methods \textit{RecoBundles}, \textit{TractQuerier} and \textit{SLAtlas}.  Methods depending on reference data (all three methods) were provided with our HCP reference data. All subjects (except for subjects from HCP datasets which did already receive basic preprocessing) were denoised (using MRtrix \citep{veraart_denoising_2016}), corrected for eddy currents and motion artifacts (using FSL eddy \citep{andersson_integrated_2016}) and rigidly registered to MNI space. This rigid registration is not required for our proposed method to work. It only requires that the left/right, front/back and up/down orientation of the images are the same as for the HCP data (i.e. images are not mirrored). Rigid registration to MNI space is an easy way to ensure this. \\
	Our proposed method showed anatomically plausible results for all subjects and most of the tracts. Only the CA was not completely reconstructed in around 20\% of the subjects. We observed partly incomplete manual reference dissections in these areas as well, indicating that the size of this very thin tract is reaching the resolution limit of the underlying imaging acquisition. Figs. \ref{fig:qualitative_results_big_ventricle}, \ref{fig:qualitative_results_MS} and \ref{fig:qualitative_results_atrophy} show exemplary results for three subjects with pathologies. For those subjects we show the corticospinal tract (CST), the optic radiation (OR) and the thalamo-postcentral tract (TPOSTC) as those tracts are heavily affected by the respective pathology. The other 19 subjects can be found in the supplementary materials.
	
	Fig. \ref{fig:qualitative_results_big_ventricle} shows the results for an alzheimer patient with abnormally large ventricles from the OASIS dataset. Even though our proposed method has only seen healthy subjects with normally sized ventricles during training it managed to properly reconstruct the CST and the OR which are heavily distorted by the enlarged ventricles. \textit{RecoBundles} also managed to find the distorted tracts. However, it failed to find the lateral projections of the CST and oversegmented the Meyer's loop of the OR. \textit{TractQuerier} showed severe oversegmentation of both the CST and OR. \textit{SLAtlas} did not manage to adapt to the enlarged ventricles. It placed the tracts inside of the ventricles as the affine registration is not able to resolve these distortions.
	
	
	Fig. \ref{fig:qualitative_results_MS} shows the results for an multiple sclerosis (MS) patient with severe lesions in the pathways of the CST and OR (marked with arrows in the figure). Inside of MS lesions demyelination takes place, leading to a loss in diffusion-weighted signal. However, the axons themselves are still intact. Therefore fibers are still running through the lesions but they are harder to reconstruct as the signal is weakened by the demyelination. Our proposed method manages to properly reconstruct streamlines running through these lesions. \textit{RecoBundles} and \textit{TractQuerier} were also able to reconstruct streamlines running through the lesions as they use tracking based on constrained spherical deconvolution which shows good results in reconstructing orientation information inside of the lesions (using a simple tensor model would not be sufficient to reconstruct the orientation information inside of the lesions). However, \textit{RecoBundles} and \textit{TractQuerier} are failing in properly reconstructing the entirety of the tracts: \textit{RecoBundles} fails to reconstruct the Meyer's loop of the OR and \textit{TractQuerier} show severe oversegmentation of both tracts. We do not show a reference tract delineation for this subject as the lesions would not be visible anymore then.
	
	
	Fig. \ref{fig:qualitative_results_atrophy} shows the results for a patient with mild brain volume loss and schizoaffective disorder. Around the postcentral gyrus the volume loss is more severe. We show results for the Thalamo-postcentral tract (T\_POSTC), containing fibers running from the thalamus to the postcentral gyrus. Despite the reduced brain volume our proposed method managed to correctly reconstruct the fibers in the postcentral gyrus. \textit{RecoBundles} is missing major parts of the tract and \textit{SLAtlas} is not able to adapt to the reduced brain volume leading to streamlines running outside of the postcentral gyrus.
	
	
	We also tested our method on subjects with brain tumors. However, given the vast distortions a tumor can produce, it is unclear where exactly certain tracts run. Experts can only assess if a tract could be a plausible reconstruction not containing any obvious errors (e.g. running through the tumor). The tract reconstructions in tumor patients produced by our method were rated as plausible by an expert. However, given the difficulty of evaluating tracts in tumor cases we do not show any results.
	
	\subsection{Runtime}
\begin{figure*}[!t]
		\centering
		\includegraphics[width=10cm]{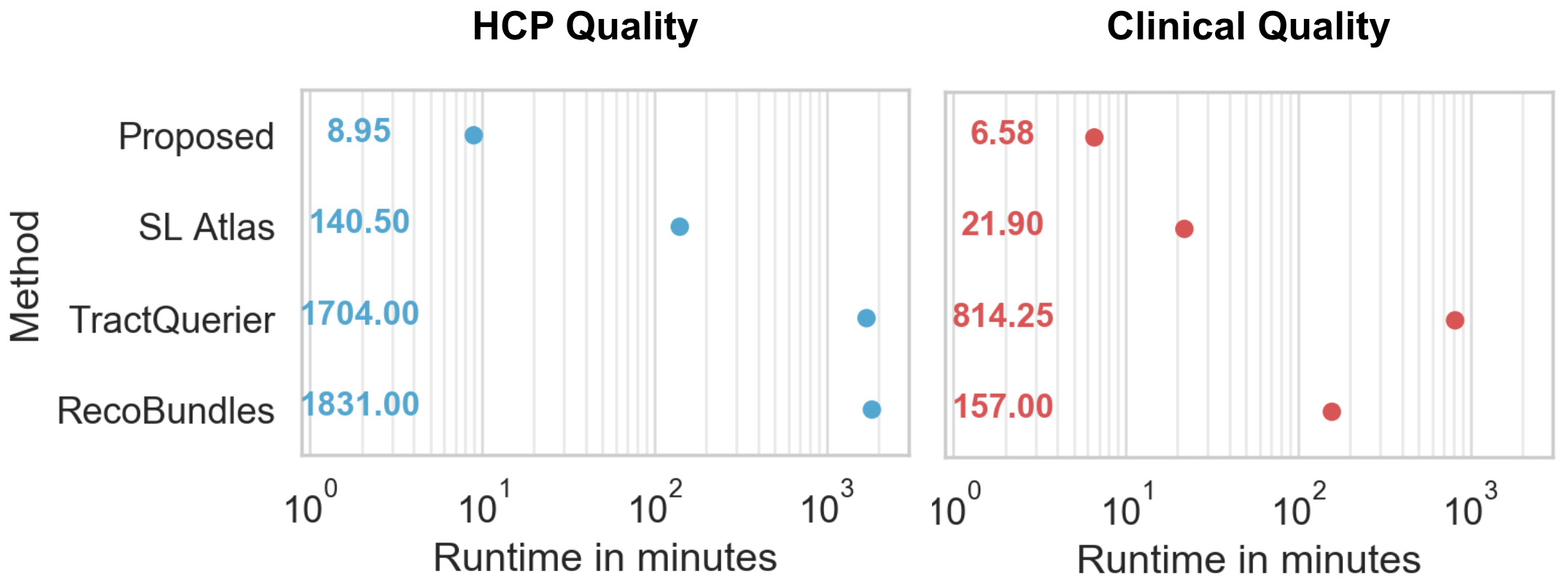}
		\caption{Runtime of our proposed method and reference methods to generate bundle-specific trackings for all 72 tracts in one subject. On \textit{HCP Quality} 10 million streamlines are used for the whole brain tracking, on \textit{Clinical Quality} only 500,000 streamlines.}
		\label{fig:runtime}
	\end{figure*}

	Runtime experiments were performed using a server with 16 2GHz Intel Xeon cores and an NVIDIA Titan X for the GPU-based approaches. We evaluated the runtime of all methods producing streamline output (\textit{Proposed}, \textit{RecoBundles}, \textit{TractQuerier} and \textit{SLAtlas}). The runtime does not include the fitting of the constrained spherical deconvolution model as this is identical for all methods. For the \textit{HCP Quality} experiments whole brain tractograms with 10 million streamlines where used. For the \textit{Clinical Quality} experiments whole brain tractograms with 500,000 streamlines were used. This reduced the runtime significantly. For our proposed approach on \textit{HCP Quality} and \textit{Clinical Quality} 2000 streamlines where generated for each tract. 
	Fig \ref{fig:runtime} shows the results for each method when reconstructing all 72 tracts in a previously unseen subject. Our method was 137x faster than the reference methods for \textit{HCP Quality} and 50x faster for \textit{Clinical Quality}.
	
	\section{Discussion and conclusion}
	
	\subsection{Overview}
	Our proposed approach is a novel method for bundle-specific tractography. It was evaluated on 72 tracts in a cohort of 105 HCP subjects in original high quality and also on reduced quality, more clinical-like datasets. Moreover we evaluated the approach on synthetic software phantoms. Seven methods were used as a benchmark. Our experiments demonstrated that our approach achieves yet unprecedented accuracy and runtime while being less affected by the reduction in resolution in the clinical quality data. It also generalizes well to unseen datasets and pathologies.
	
	\subsection{Reference data}
	The tract delineations from the reference dataset used for training and evaluation do not represent a real ground truth. They are approximations based on diffusion weighted images, which has several limitations. However, given the high quality of the HCP data and the manual inspection by an expert, the employed dataset represents one of the best existing in vivo approximations of known white matter anatomy in a cohort of that size. Moreover, by using synthetic software phantoms we were able to evaluate our method on a dataset where the real ground truth is available.
	
	On the \textit{Phantom} data Dice scores were lower than on the \textit{Clinical Quality} data.
	This had two main reasons: First the phantoms were simulated containing major artifacts whereas \textit{Clinical Quality} contains only little artifacts as it based one the high quality HCP data. Second the domain shift between the training data and the \textit{Phantom} data is significantly greater than the shift between the training data and the \textit{Clinical Quality} data: On the one hand different acquisition settings were used during phantom simulation and on the other hand the simulated brains only contain the 72 reference tracts. Those cover the majority of the brain, but several tracts (like for example all u-fibers) were not included in the phantom. Therefore the training data and the phantoms are less similar and the resulting scores are reduced.\\
	Despite the reduced Dice scores in comparison to the \textit{Clinical Quality} data, our approach still shows complete reconstructions on the \textit{Phantom} dataset with only minor errors (Fig. \ref{fig:qualitative_results_phantom} in supplementary materials).
	
	\subsection{Reference methods}
	Selecting appropriate reference methods for a fair comparison was not easy as all methods have slightly different approaches and requirements. The comparison with our selected reference methods is also subject to some limitations:
	\textit{TractQuerier} was part of the pipeline used for the creation of the reference dataset which was then used to evaluate \textit{TractQuerier} against, thus inducing a potential positive bias for the method.
	For \textit{RecoBundles} we were only able to use 5 reference subjects due to the long runtime of \textit{RecoBundles}. Using all 63 subjects from the training set as reference subjects would have been computationally infeasible for 72 tracts and tractograms with 10 million streamlines. Moreover, as suggested by our \textit{Atlas Custom} and \textit{Multi-Mask} experiments, averaging more subjects, does not necessarily increase accuracy as small details become blurred. Using 5 reference subjects therefore provides a good estimation of the performance of \textit{RecoBundles}. We used the default \textit{RecoBundles} settings. Optimizing those might improve the results to some degree.
	\textit{SLAtlas} is showing high angular errors because it is only based on affine registration making the registered tract not align properly on the new subject. \textit{Peak Atlas} is based on elastic registration which leads to better alignment of tracts. However, it is also showing higher angular errors as elastic registration is still not able to completely resolve the complex inter-subject variability that exists between human brains.
	
	As we have shown, our comparison to the reference methods has some limitations. However, those limitations do not apply to all reference methods and those limitations alone cannot explain the large accuracy gap between our method and all reference methods, indicating the great potential of the proposed method.
	
	\subsection{Preprocessing}
	The preprocessing of the reference data to extract start/end region segmentations and to determine the main streamline orientation in each voxel is made up of several non-trivial steps (see section \ref{sec:data_processing}) containing a multitude of parameters which have to be chosen. The influence of these parameters on the final results was not systematically evaluated. Therefore changes to the parameters could significantly alter the final results.
	
	\subsection{Generalization}
	Our method is based on supervised learning, bearing the inherent limitation of depending on the availability and quality of training data. This is similar to most of the reference methods which also require reference tracts or atlases. Using scanners and acquisition sequences different from the training data introduces a domain shift and therefore reduces the performance. By using heavy data augmentation during training this domain shift can be reduced. Our experiments on the phantom data have shown that our method generalizes well to unseen acquisition sequences. We have also shown on a wide range of unseen datasets from different scanners with and without pathologies that the proposed method produces anatomically plausible results in most cases.
	
	\subsection{Code availability}
	The proposed method is openly available as an easy-to-use python package with pretrained weights: https://github.com/MIC-DKFZ/TractSeg/

	\section*{Acknowledgments}
	HCP data were provided by the Human Connectome Project, WU- Minn Consortium (Principal Investigators: David Van Essen and Kamil Ugurbil; 1U54MH091657) funded by the 16 NIH Institutes and Centers that support the NIH Blueprint for Neuroscience Research; and by the McDonnell Center for Systems Neuroscience at Washington University.
	Data used in preparation for this article were obtained from the SchizConnect database (http://schizconnect.org). As such, the investigators within SchizConnect contributed to the design and implementation of SchizConnect and/or provided data but did not participate in analysis or writing of this report. Data collection and sharing for this project was funded by NIMH cooperative agreement 1U01 MH097435.
	BrainGluSchi data was downloaded from the COllaborative Informatics and Neuroimaging Suite Data Exchange tool (COINS; http://coins.mrn.org/dx) and data collection was funded by NIMH R01MH084898- 01A1.
	COBRE data was downloaded from the COllaborative Informatics and Neuroimaging Suite Data Exchange tool (COINS; http://coins.mrn.org/dx) and data collection was performed at the Mind Research Network, and funded by a Center of Biomedical Research Excellence (COBRE) grant 5P20RR021938/P20GM103472 from the NIH to Dr. Vince Calhoun.
	This work was supported by the German Research Foundation (DFG) grant MA 6340/10-1 and grant MA 6340/12-1.
	Data collection and sharing for this project was funded by the Alzheimer's Disease Neuroimaging Initiative
	(ADNI) (National Institutes of Health Grant U01 AG024904) and DOD ADNI (Department of Defense award
	number W81XWH-12-2-0012). ADNI is funded by the National Institute on Aging, the National Institute of
	Biomedical Imaging and Bioengineering, and through generous contributions from the following: AbbVie,
	Alzheimer’s Association; Alzheimer’s Drug Discovery Foundation; Araclon Biotech; BioClinica, Inc.; Biogen;
	Bristol-Myers Squibb Company; CereSpir, Inc.; Cogstate; Eisai Inc.; Elan Pharmaceuticals, Inc.; Eli Lilly and
	Company; EuroImmun; F. Hoffmann-La Roche Ltd and its affiliated company Genentech, Inc.; Fujirebio; GE
	Healthcare; IXICO Ltd.; Janssen Alzheimer Immunotherapy Research \& Development, LLC.; Johnson \&
	Johnson Pharmaceutical Research \& Development LLC.; Lumosity; Lundbeck; Merck \& Co., Inc.; Meso
	Scale Diagnostics, LLC.; NeuroRx Research; Neurotrack Technologies; Novartis Pharmaceuticals
	Corporation; Pfizer Inc.; Piramal Imaging; Servier; Takeda Pharmaceutical Company; and Transition
	Therapeutics. The Canadian Institutes of Health Research is providing funds to support ADNI clinical sites
	in Canada. Private sector contributions are facilitated by the Foundation for the National Institutes of Health
	(www.fnih.org). The grantee organization is the Northern California Institute for Research and Education,
	and the study is coordinated by the Alzheimer’s Therapeutic Research Institute at the University of Southern
	California. ADNI data are disseminated by the Laboratory for Neuro Imaging at the University of Southern
	California.
	Data [in part] was collected at Brain and Mind centre, Sydney University and funded by NMSS and Novartis.
	Data were provided [in part] by OASIS-3: Principal Investigators: T. Benzinger, D. Marcus, J. Morris; NIH P50AG00561, P30NS09857781, P01AG026276, P01AG003991, R01AG043434, UL1TR000448, R01EB009352. AV-45 doses were provided by Avid Radiopharmaceuticals, a wholly owned subsidiary of Eli Lilly.

	\section*{Supplementary materials}

	\subsection*{List of all tracts}
	\vbox{%
	This is a list of all 72 tracts supported by our model: Arcuate fascicle (AF), Anterior thalamic radiation (ATR), Anterior commissure (CA), Corpus callosum (Rostrum (CC 1), Genu (CC 2), Rostral body (CC 3), Anterior midbody (CC 4), Posterior midbody (CC 5), Isthmus (CC 6), Splenium (CC 7)), Cingulum (CG), Corticospinal tract (CST), Middle longitudinal fascicle (MLF), Fronto-pontine tract (FPT), Fornix (FX), Inferior cerebellar peduncle (ICP), Inferior occipito-frontal fascicle (IFO), Inferior longitudinal fascicle (ILF), Middle cerebellar peduncle (MCP), Optic radiation (OR), Parieto-occipital pontine (POPT), Superior cerebellar peduncle (SCP), Superior longitudinal fascicle I (SLF I), Superior longitudinal fascicle II (SLF II), Superior longitudinal fascicle III (SLF III), Superior thalamic radiation (STR), Uncinate fascicle (UF), Thalamo-prefrontal (T\_PREF), Thalamo-premotor (T\_PREM), Thalamo-precentral (T\_PREC), Thalamo-postcentral (T\_POSTC), Thalamo-parietal (T\_PAR), Thalamo-occipital (T\_OCC), Striato-fronto-orbital (ST\_FO), Striato-prefrontal (ST\_PREF), Striato-premotor (ST\_PREM), Striato-precentral (ST\_PREC), Striato-postcentral (ST\_POSTC), Striato-parietal (ST\_PAR), Striato-occipital (ST\_OCC)
	}

	\subsection*{RecoBundles default parameters}
    \vbox{%
	Parameters for whole brain streamline-based registration:
	\begin{itemize}
	    \item streamline-based linear registration transform: affine
	    \item streamline-based linear registration progressive: True
	    \item maximum iterations for registration optimization: 150
	    \item random streamlines for starting QuickBundles: 50000
	\end{itemize}
	
	Parameters for recognize bundles:
	\begin{itemize}
    	\item clustering threshold: 15
    	\item reduction threshold: 10
    	\item reduction distance: mdf
    	\item model clustering threshold: 5
    	\item pruning threshold: 5
    	\item pruning distance: mdf
    	\item local streamline-based linear registration: True
    	\item streamline-based linear registration metric: None
    	\item streamline-based linear registration transform: similarity
    	\item streamline-based linear registration progressive: True
    	\item streamline-based linear registration matrix: small
	\end{itemize}
    }
	
	\begin{table*}[!t]
		\small
		\caption{Acquisition parameters of additional test datasets.}
		\label{tab:datasets}
		\centering
		\begin{tabular}{|l|p{2.3cm}|p{2.1cm}|p{2.8cm}|p{1.1cm}|}
			\hline
			\textbf{Project} & \textbf{Pathology} & \textbf{Resolution$^*$}& \textbf{b-Values} & \textbf{Field strength} \\
			\hline
			TRACED $^1$ & healthy & 2.5mm & 3x $b=0mm/s^2$\newline
			20x $b=1000mm/s^2$ \newline 48x $b=2000mm/s^2$ \newline
			64x $b=3000mm/s^2$ & 3T\\
			\hline
			Internal (Healthy) & healthy & 2.5mm & 1x $b=0mm/s^2$\newline
			81x $b=1000mm/s^2$ \newline 81x $b=2000mm/s^2$ \newline
			81x $b=3000mm/s^2$ & 3T\\
			\hline
			BrainGluShi \citep{bustillo_glutamatergic_2017} $^2$ & healthy & 2.0mm & 5x $b=0mm/s^2$\newline
			30x $b=800mm/s^2$ & 3T\\
			\hline		
			Stanford\_hardi \citep{rokem_high_2013} $^3$ & healthy & 2.0mm & 10x $b=0mm/s^2$\newline
			150x $b=2000mm/s^2$ & 3T\\
			\hline		
			Sherbrooke\_3shell $^4$ & healthy & 2.5mm & 1x $b=0mm/s^2$\newline
			64x $b=1000mm/s^2$ \newline 64x $b=2000mm/s^2$ \newline
			64x $b=3500mm/s^2$ & 3T\\
			\hline
			Rockland \citep{nooner_nki-rockland_2012} $^5$ & healthy & 2.0mm & 9x $b=0mm/s^2$\newline
			128x $b=1500mm/s^2$ & 3T\\
			\hline	
			HCP 7T \citep{van_essen_wu-minn_2013} & healthy & 1.05mm & 15 $b=0mm/s^2$\newline
			64x $b=1000mm/s^2$  \newline 64x $b=2000mm/s^2$& 7T\\
			\hline
			IXI $^6$ & healthy $>$ 80 years & 1.75x1.75x2mm & 1 $b=0mm/s^2$\newline
			15x $b=1000mm/s^2$ & 3T\\
			\hline	
			HCP lifespan \citep{tisdall_volumetric_2012} $^7$ & healthy $<$ 10 years & 1.5mm & 10 $b=0mm/s^2$\newline
			76x $b=1000mm/s^2$  \newline 75x $b=2500mm/s^2$& 3T\\
			\hline	
			COBRE \citep{cetin_thalamus_2014} $^2$ & schizophrenia, enlarged ventricles & 2.0mm & 5x $b=0mm/s^2$\newline 30x $b=800mm/s^2$ & 3T\\
			\hline		
			SoftSigns \citep{hirjak_white_2017} & neurological soft signs & 2.5mm & 1x $b=0mm/s^2$\newline 81x $b=1000mm/s^2$ & 3T\\
			\hline
			Internal (Autism) & autism spectrum disorder & 2.5mm & 5x $b=0mm/s^2$\newline 60x $b=1000mm/s^2$ & 3T\\
			\hline				
			Internal (Schizophrenia) & schizophrenia & 1.7mm & 3x $b=0mm/s^2$\newline 60x $b=1500mm/s^2$ & 3T\\
			\hline				
			CNP \citep{poldrack_phenome-wide_2016} $^8$ & schizophrenia, bipolar, ADHD & 2.0mm & 1x $b=0mm/s^2$\newline 64x $b=1000mm/s^2$ & 3T\\
			\hline
			Internal (MS) & multiple sclerosis & 1x1x2mm & 2x $b=0mm/s^2$\newline 64x $b=1000mm/s^2$ & 3T\\
			\hline
			ADNI $^9$ & alzheimer & 1.4x1.4x2.7mm & 5x $b=0mm/s^2$\newline 41x $b=1000mm/s^2$ & 3T\\
			\hline
			OASIS $^{10}$ & alzheimer & 2.5mm & 1x $b=0mm/s^2$\newline 64x $b=1000mm/s^2$ & 3T\\
			\hline
			
			\multicolumn{5}{@{}l}{\parbox{14cm}{$^1$https://my.vanderbilt.edu/ismrmtraced2017/ \newline
					$^2$ http://schizconnect.org/ \newline
					$^3$ https://purl.stanford.edu/yx282xq2090 \newline
					$^4$ http://nipy.org/dipy/reference/dipy.data.html\#fetch-sherbrooke-3shell \newline
					$^5$ http://fcon\_1000.projects.nitrc.org/indi/enhanced/ \newline
					$^6$ https://brain-development.org/ixi-dataset/ \newline
					$^7$ https://www.humanconnectome.org/study-hcp-lifespan-pilot \newline
					$^8$ https://openneuro.org/datasets/ds000030/versions/00016 \newline
					$^9$ http://adni.loni.usc.edu/data-samples/access-data/ \newline
					$^{10}$ http://www.oasis-brains.org \newline
					$^*$ if only one value is shown, the resolution is isotropic}}
		\end{tabular}
	\end{table*}
	
	\begin{figure*}[!t]
		\centering
		\includegraphics[width=\textwidth]{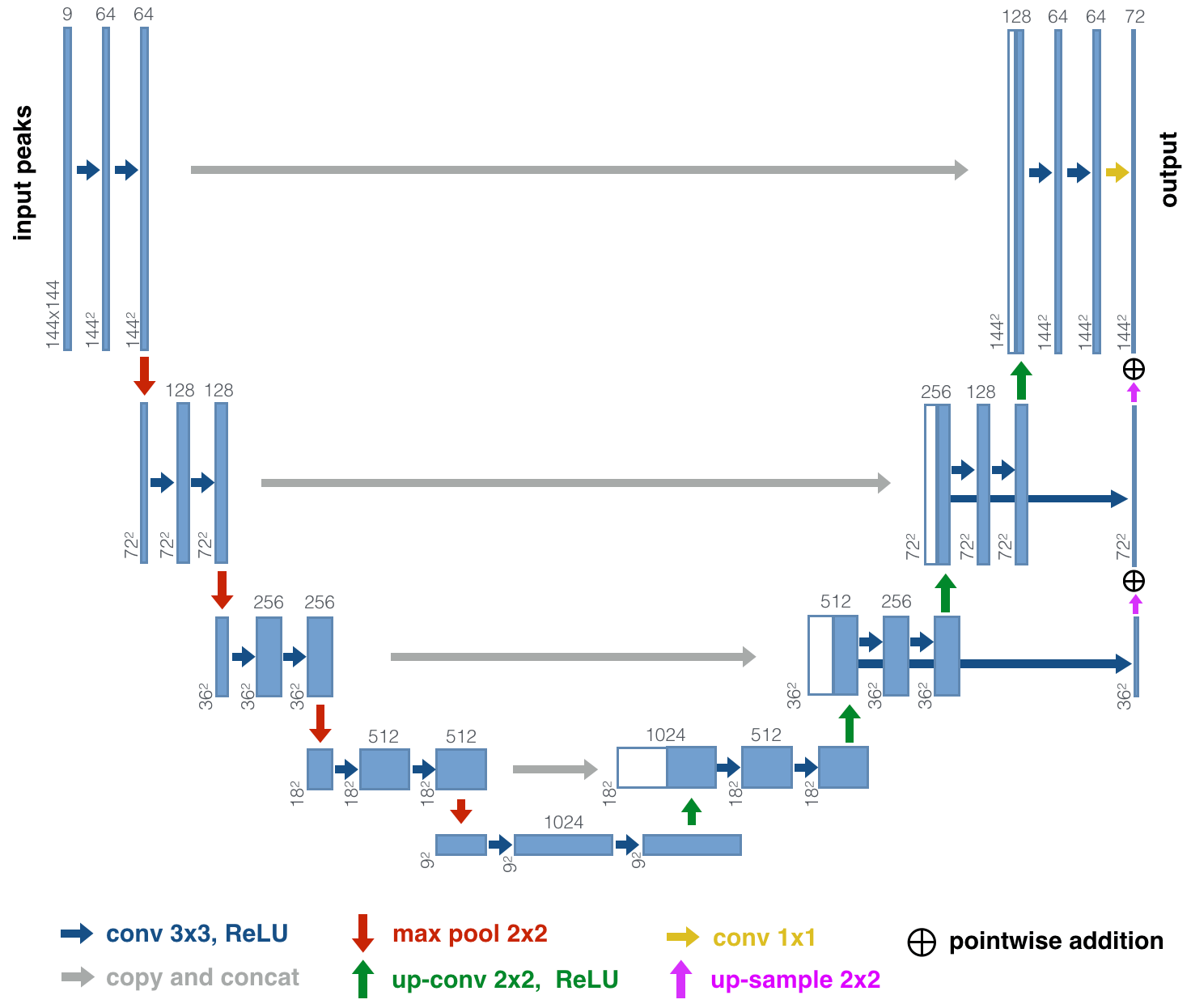}
		\caption{Proposed U-Net architecture. Blue boxes represent multi-channel feature maps. White boxes show copied feature maps. The gray number on top of each box gives the number of channels, the x-y-size is given at lower left corner of each box. Network operations are represented by differently colored arrows. The main difference to the original U-Net architecture are the extra convolutions layers in the upsampling path allowing better gradient flow (deep supervision).}
		\label{fig:unet}
	\end{figure*}

	\begin{figure*}[!t]
		\centering
		\includegraphics[width=\textwidth]{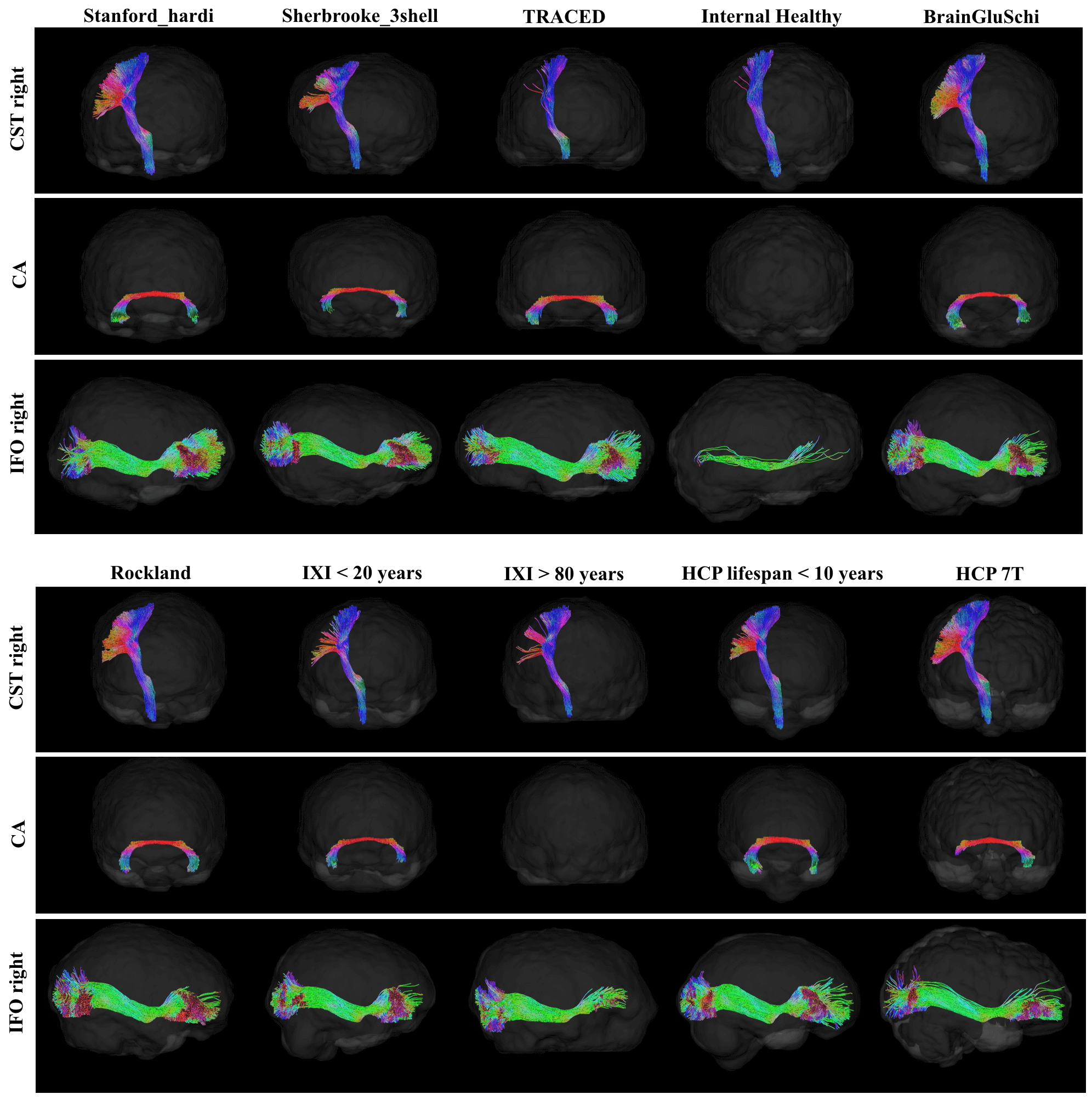}
		\caption{Qualitative results of our proposed method on 10 healthy subjects from 9 different datasets (see table \ref{tab:datasets}) while being trained on the HCP reference dataset: reconstruction of right corticospinal tract (CST), anterior commissure (CA) and right inferior occipito-frontal fascicle (IFO).}
		\label{fig:supplementary_non_hcp_healthy}
	\end{figure*}

	\begin{figure*}[!t]
		\centering
		\includegraphics[width=\textwidth]{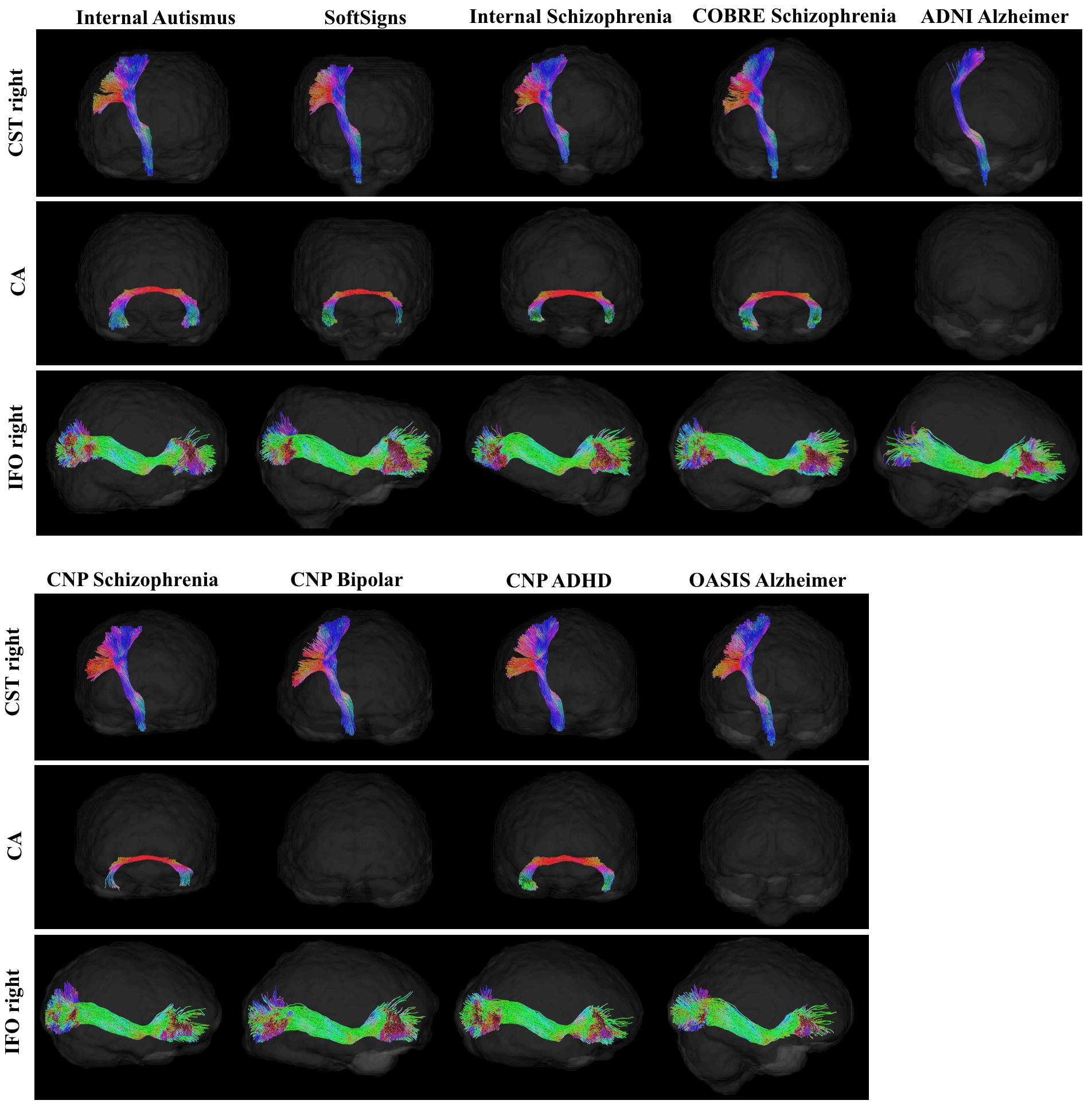}
		\caption{Qualitative results of our proposed method on 9 subjects with pathologies from 7 different datasets (see table \ref{tab:datasets}) while being trained on the HCP reference dataset: reconstruction of right corticospinal tract (CST), anterior commissure (CA) and right inferior occipito-frontal fascicle (IFO).}
		\label{fig:supplementary_non_hcp_pathology}
	\end{figure*}

	\begin{figure*}[!t]
		\centering
		\includegraphics[width=\textwidth]{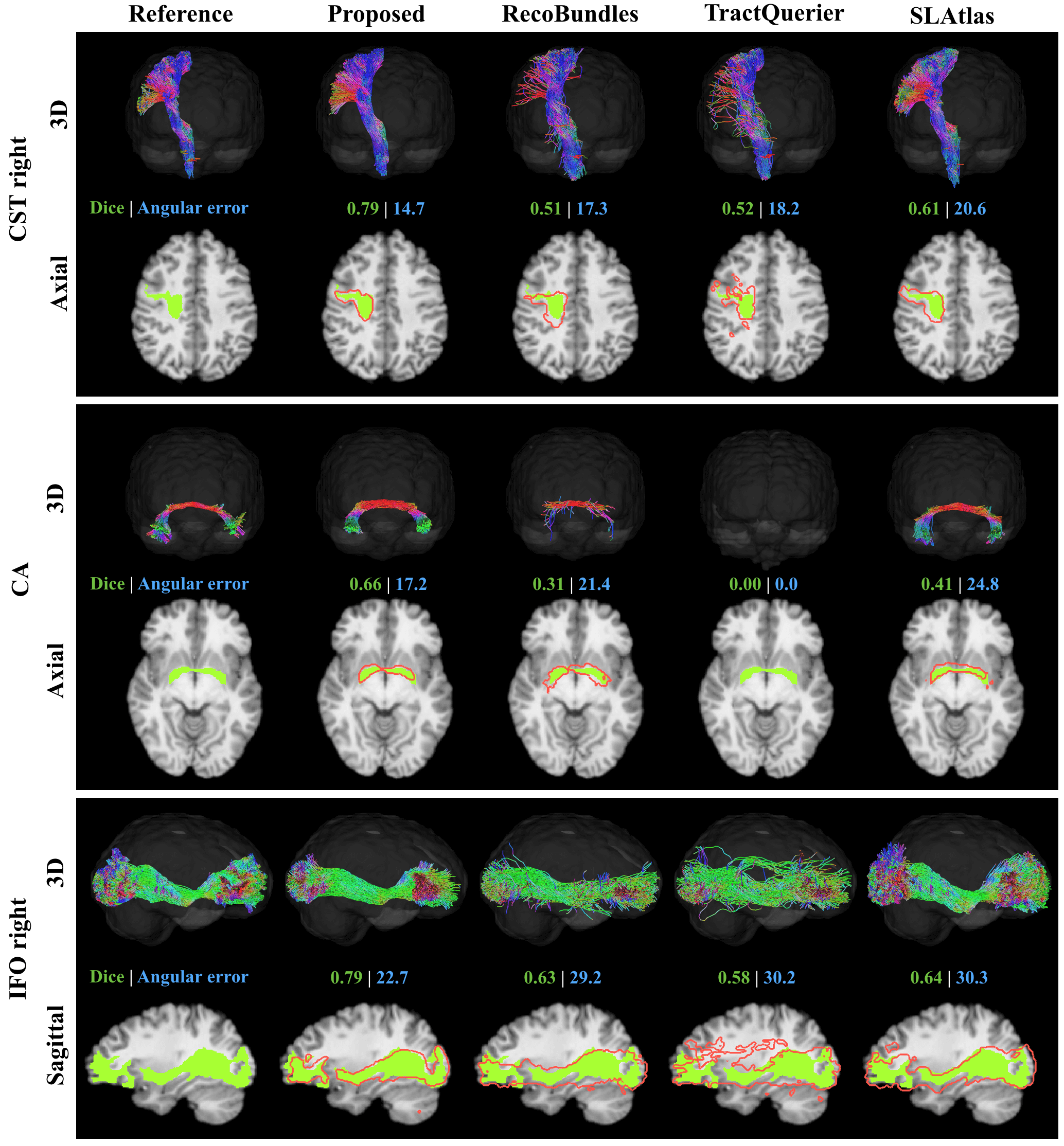}
		\caption{Qualitative comparison of results on \textit{Phantom} test set: reconstruction of right corticospinal tract (CST), anterior commissure (CA) and right inferior occipito-frontal fascicle (IFO) on subject 623844. Green shows the reference tract and red shows the tract mask of the respective method.}
		\label{fig:qualitative_results_phantom}
	\end{figure*}

	\begin{figure*}[!t]
		\centering
		\includegraphics[width=0.8\linewidth]{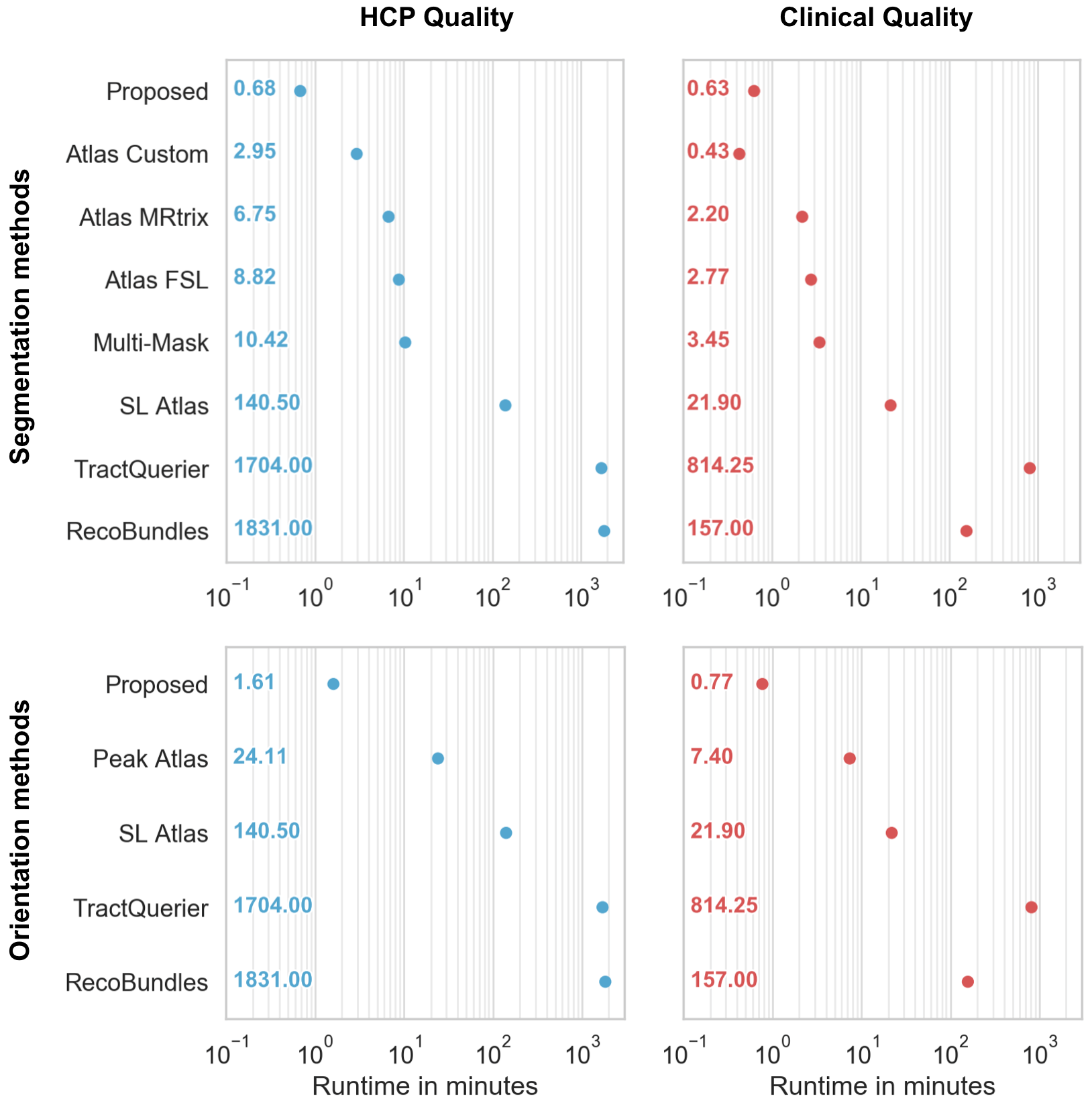}
		\caption{Runtime (in minutes) of all methods for segmentation and for estimation of orientation shown in the quantitative evaluation.}
		\label{fig:runtime_all_methods}
	\end{figure*}

	\begin{figure*}[!t]
		\centering
		\includegraphics[width=0.7\linewidth]{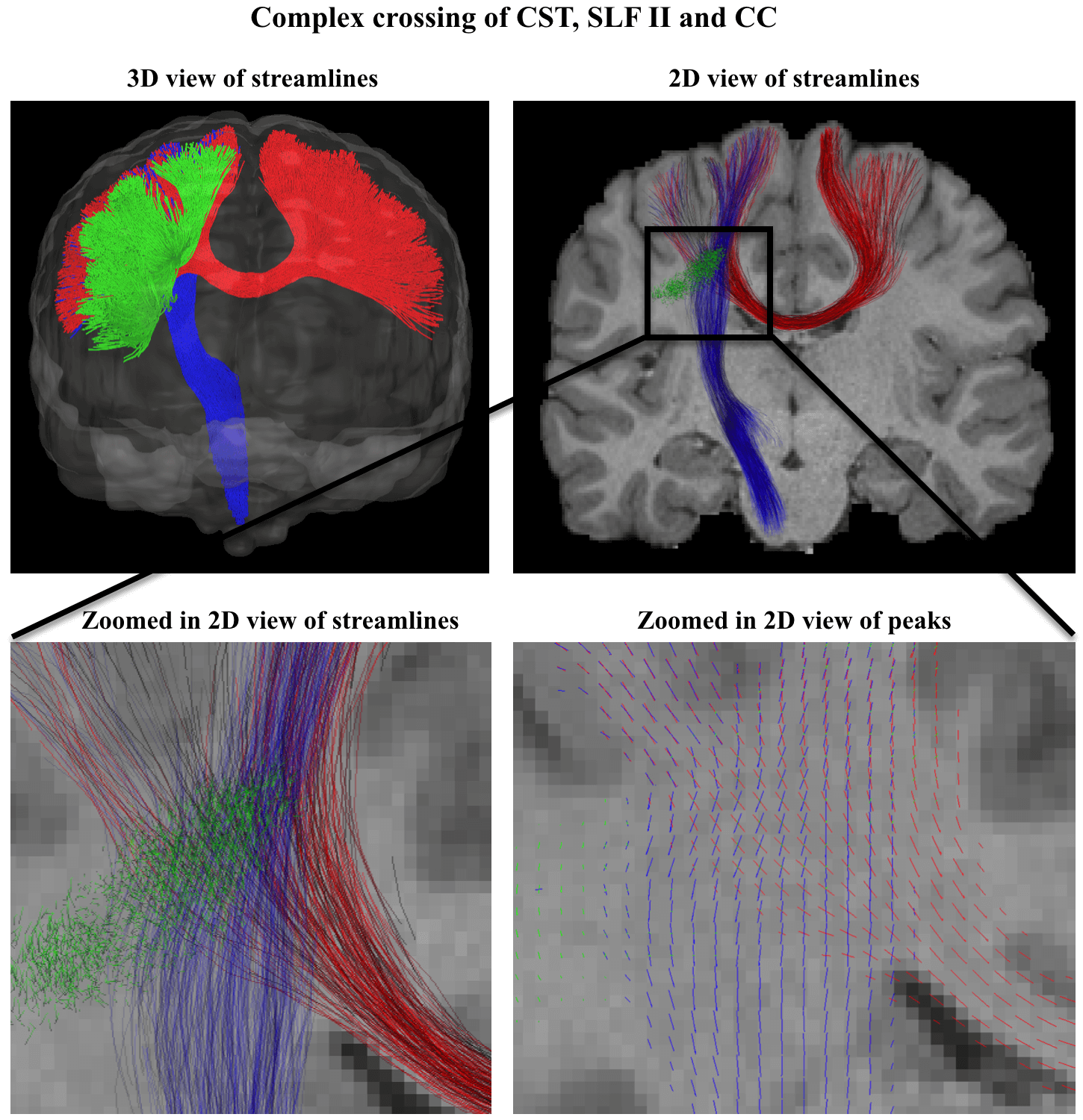}
		\caption{The figure shows the results for the proposed method on the crossing of CST (blue), SLF II (green) and CC 4 (blue) on subject 623844 of the \textit{HCP Quality} dataset. The zoomed in peak image shows the TOM peaks which the tracking is based on. The proposed method manages to properly resolve this complex crossing of three major tracts.}
		\label{fig:complex_crossing}
	\end{figure*}

	\begin{figure*}[!t]
		\centering
		\includegraphics[width=\textwidth]{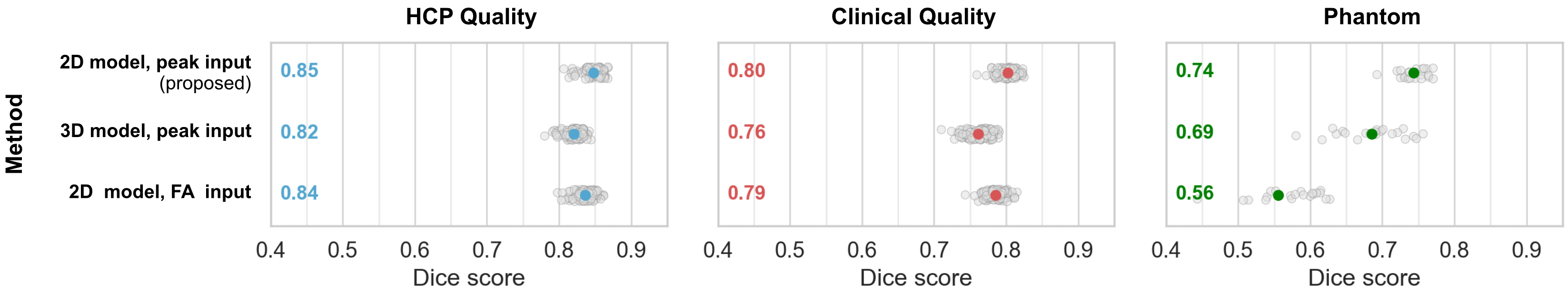}
		\caption{Comparison of segmentation results for 2D and 3D model as well as for peak input and FA input. For the 3D model to fit into the memory we had to reduce the batch size to 1, the number of filters by a factor of 8 and the number of downsampling levels from 4 to 3. The 2D model with peak input showed the best results. In terms of Dice score the differences on the \textit{HCP Quality} and \textit{Clinical Quality} dataset are only minor. But differences on the \textit{Phantom} dataset are more severe. Also when looking at results on the 17 non-HCP datasets the 3D model as well as the FA input showed clearly worse results, missing major parts of several bundles.}
		\label{fig:fa_3D}
	\end{figure*}

	\begin{figure*}[!t]
		\centering
		\includegraphics[width=0.7\linewidth]{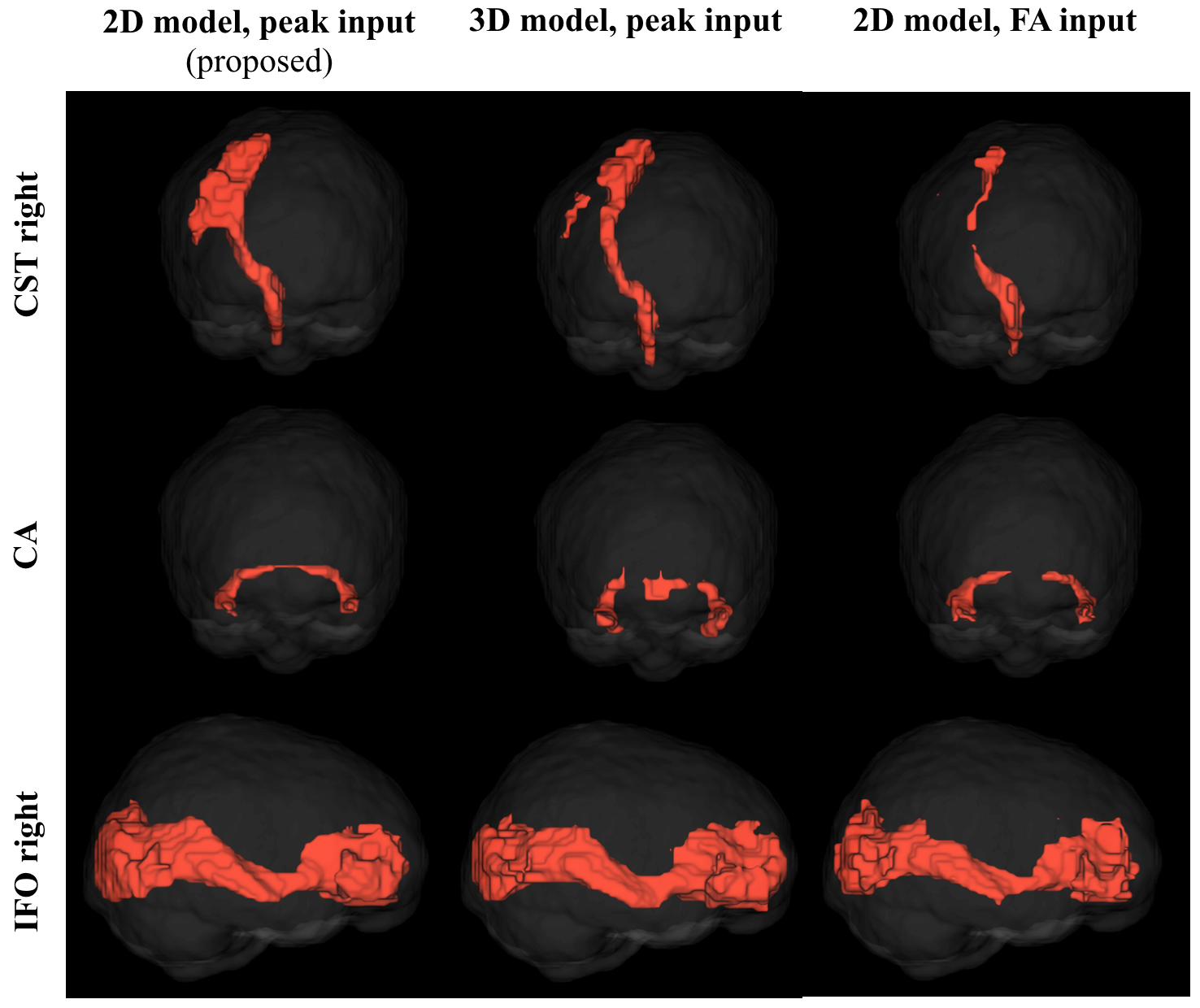}
		\caption{Segmentation results for 2D and 3D model as well as for peak input and FA input for one subject from the COBRE dataset with enlarged ventricles. The proposed method (2D model with peak input) manages to completely segment all bundles. The 3D model as well as the 2D model using FA as input are missing major parts of several bundles.}
		\label{fig:fa_3D_qualitative}
	\end{figure*}

	\begin{figure*}[!t]
		\centering
		\includegraphics[width=\textwidth]{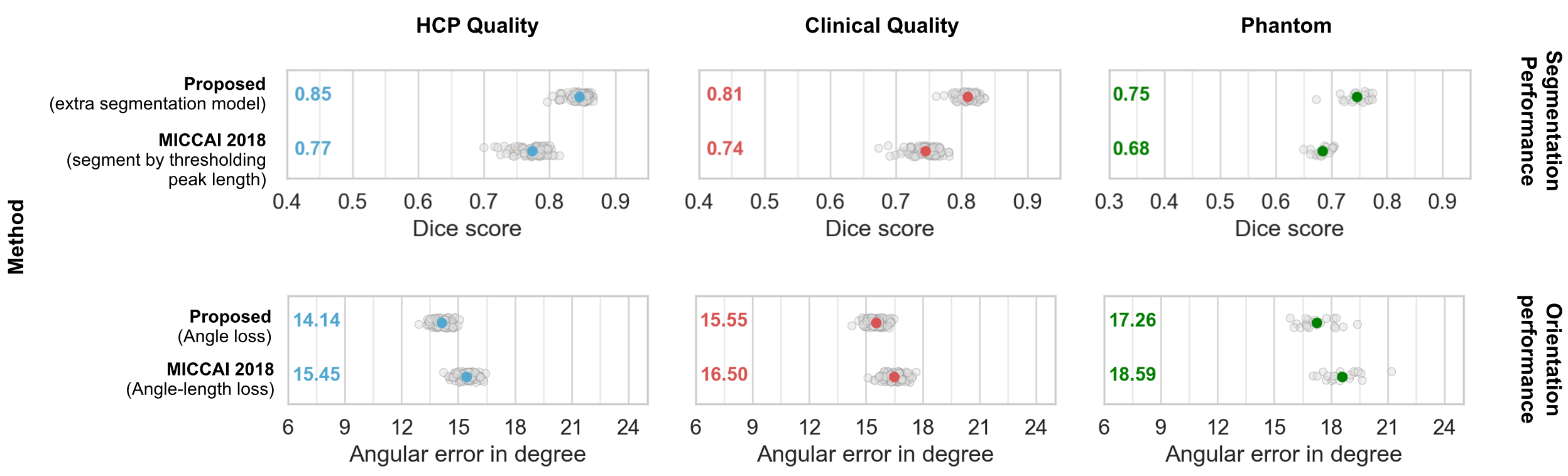}
		\caption{Comparison of the proposed method to results from Wasserthal et al. (2018a) at MICCAI 2018. By using an extra model for segmentation and an extra model for learning of the peak angle we achieved better results than using a model which tries to learn both at the same time (by learning a peak angle and a peak length).}
		\label{fig:miccai_quantitative_comparison}
	\end{figure*}

	\begin{figure*}[!t]
		\centering
		\includegraphics[width=0.7\linewidth]{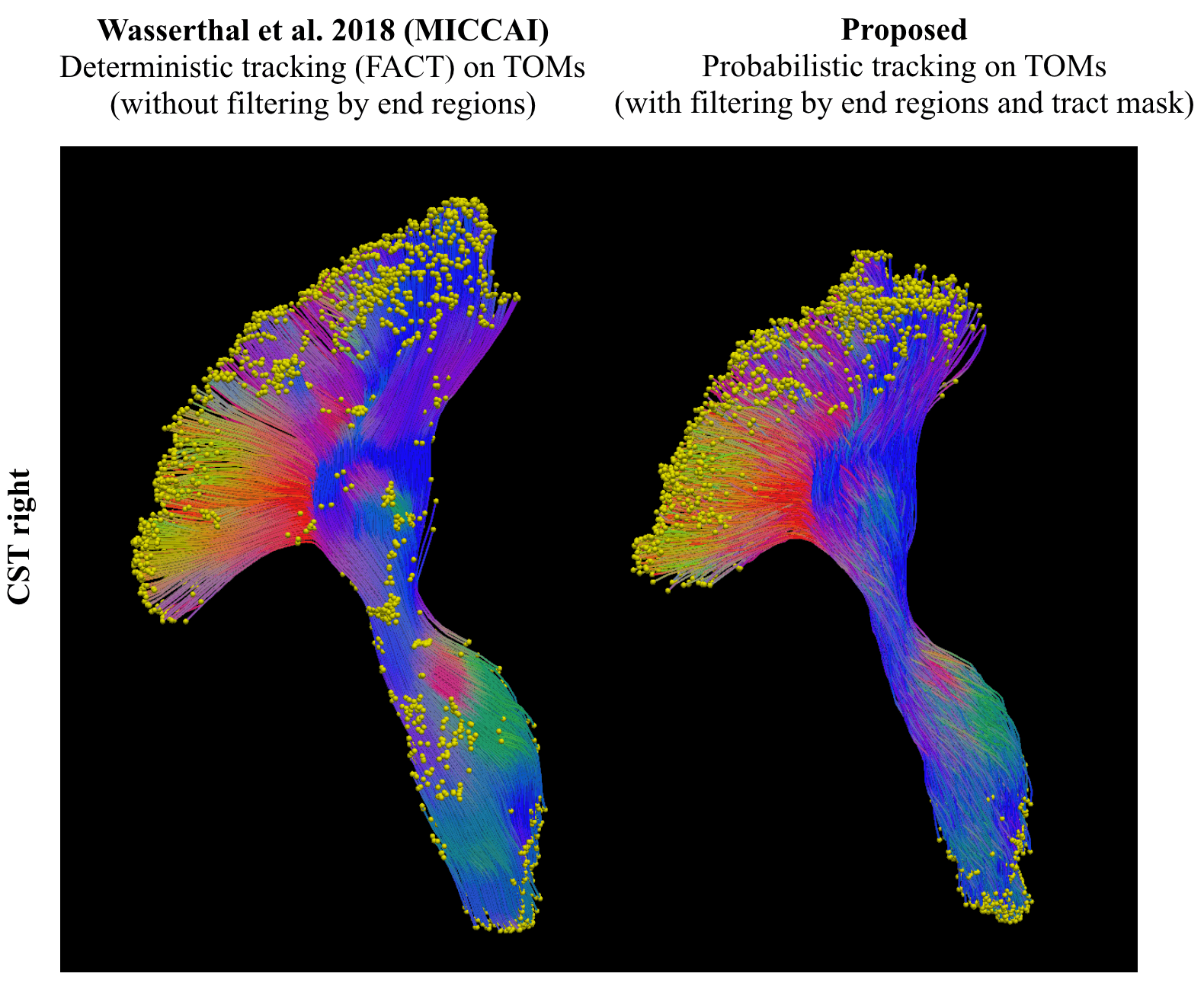}
		\caption{Comparison of the proposed method to results from Wasserthal et al. (2018a) at MICCAI 2018 for one subject from the BrainGluSchi dataset. Streamline endpoints are marked with yellow dots. By filtering streamlines by start and end region masks streamlines ending prematurely can be removed. This leads to sparser results. By using custom probabilistic tracking complete tractograms can be obtained while still filtering by endpoints.}
		\label{fig:miccai_qualitative_comparison}
	\end{figure*}

	\begin{figure*}[!t]
		\centering
		\includegraphics[width=0.9\linewidth]{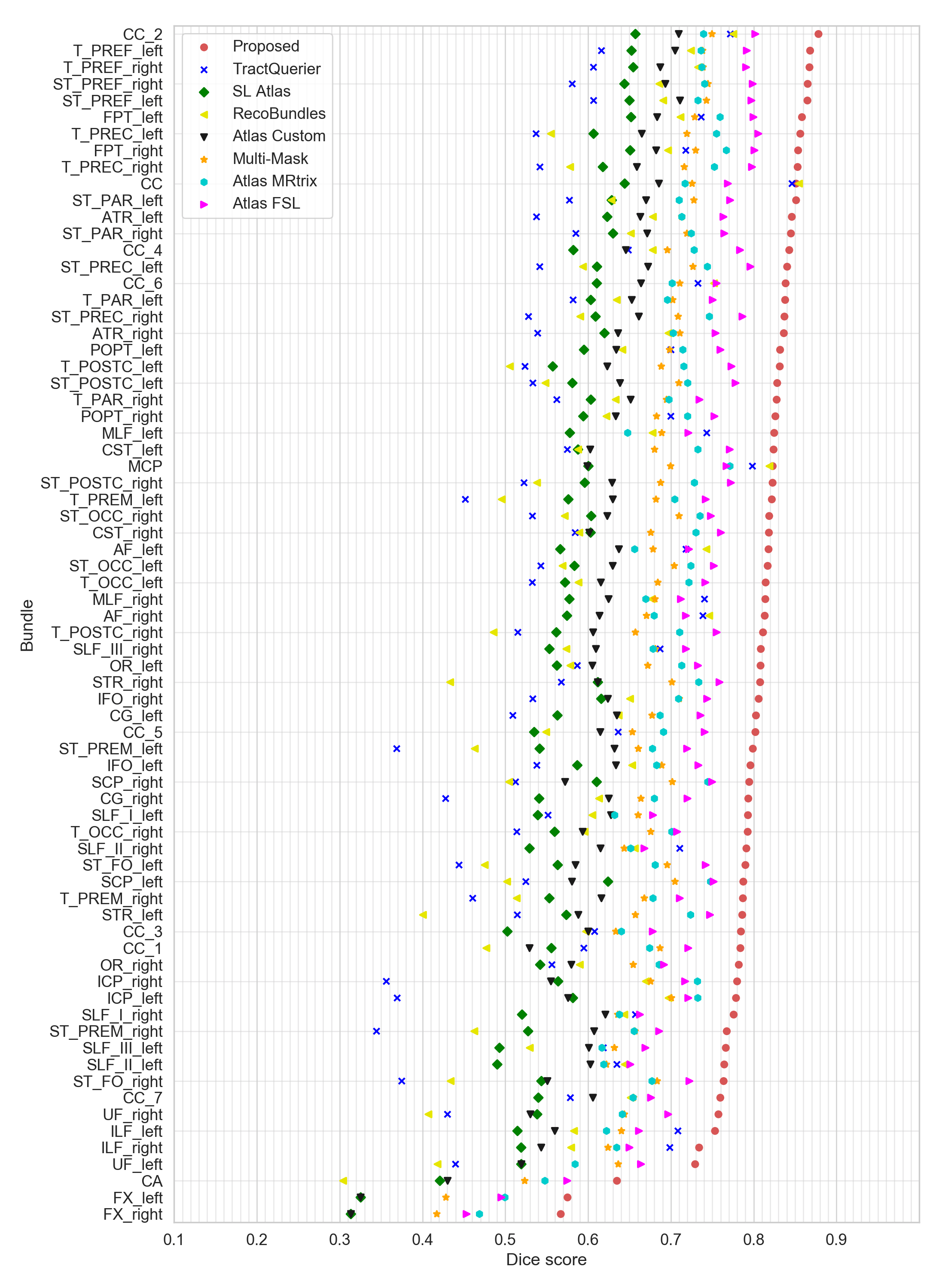}
		\caption{Dice scores for all 72 tracts on the Clinical Quality dataset for our proposed method and all reference methods sorted by score. The full name of each tract can be seen in the supplementary materials.}
		\label{fig:results_per_bundle_lowRes}
	\end{figure*}

	\begin{figure*}[!t]
		\centering
		\includegraphics[width=0.9\linewidth]{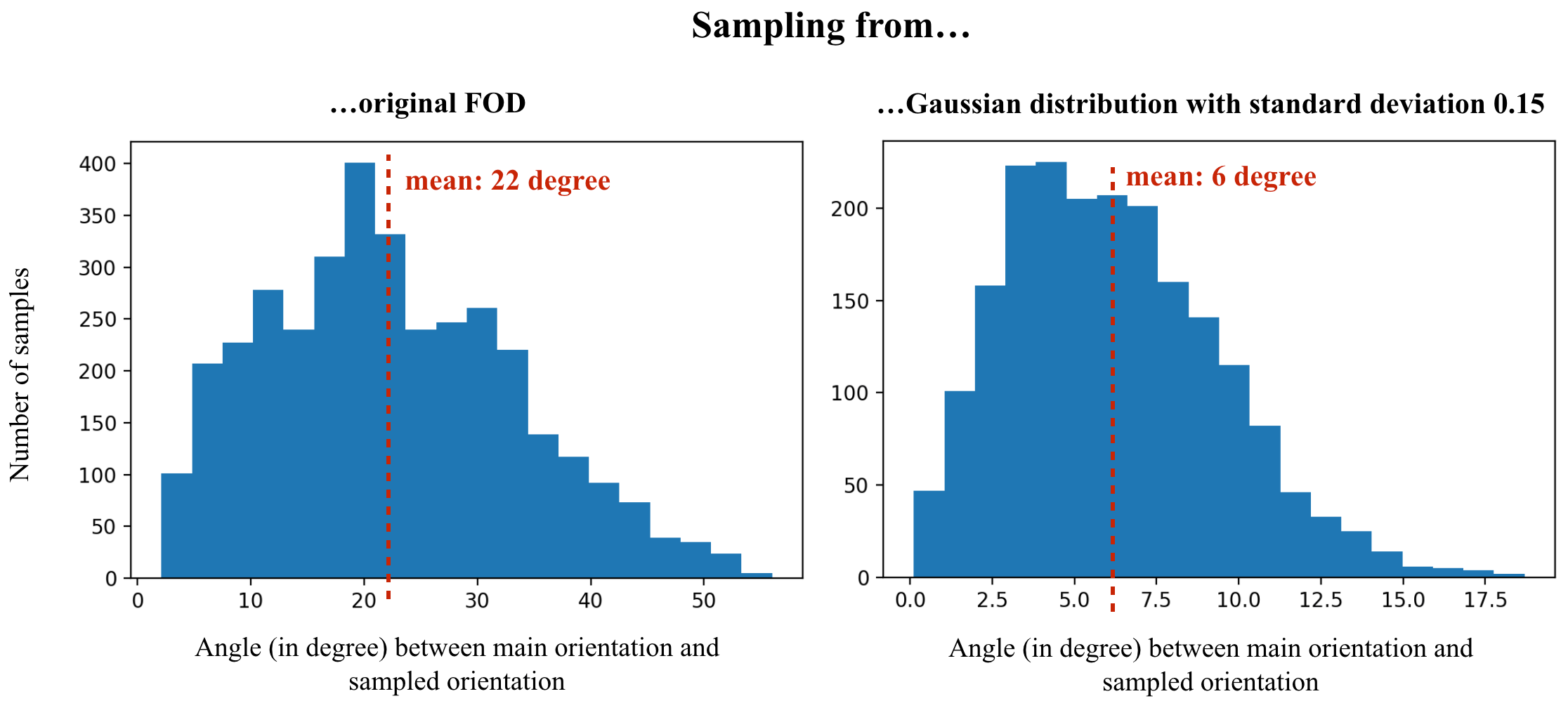}
		\caption{Histogram of sampled orientations for one voxel in the corpus callosum of one HCP subject. The x-axis shows the angle between the main orientation and the sampled orientation. On the left is the result for the original FOD (generated with constrained spherical deconvolution). On the right is the result for sampling from a fixed Gaussian distribution with standard deviation 0.15. As can be seen when using a standard deviation of 0.15 the fixed distribution has clearly lower dispersion in the angles than the original FOD, thus providing a conservative lower bound.}
		\label{fig:dispersion}
	\end{figure*}


\begin{thebibliography}{45}
\expandafter\ifx\csname natexlab\endcsname\relax\def\natexlab#1{#1}\fi
\providecommand{\url}[1]{\texttt{#1}}
\providecommand{\href}[2]{#2}
\providecommand{\path}[1]{#1}
\providecommand{\DOIprefix}{doi:}
\providecommand{\ArXivprefix}{arXiv:}
\providecommand{\URLprefix}{URL: }
\providecommand{\Pubmedprefix}{pmid:}
\providecommand{\doi}[1]{\href{http://dx.doi.org/#1}{\path{#1}}}
\providecommand{\Pubmed}[1]{\href{pmid:#1}{\path{#1}}}
\providecommand{\bibinfo}[2]{#2}
\ifx\xfnm\relax \def\xfnm[#1]{\unskip,\space#1}\fi
\bibitem[{Alexander et~al.(2001)Alexander, Pierpaoli, Basser and
  Gee}]{alexander_spatial_2001}
\bibinfo{author}{Alexander, D.C.}, \bibinfo{author}{Pierpaoli, C.},
  \bibinfo{author}{Basser, P.J.}, \bibinfo{author}{Gee, J.C.},
  \bibinfo{year}{2001}.
\newblock \bibinfo{title}{Spatial transformations of diffusion tensor magnetic
  resonance images}.
\newblock \bibinfo{journal}{IEEE transactions on medical imaging}
  \bibinfo{volume}{20}, \bibinfo{pages}{1131--1139}.
\newblock \DOIprefix\doi{10.1109/42.963816}.
\bibitem[{Alexander et~al.(2017)Alexander, Zikic, Ghosh, Tanno, Wottschel,
  Zhang, Kaden, Dyrby, Sotiropoulos, Zhang and
  Criminisi}]{alexander_image_2017}
\bibinfo{author}{Alexander, D.C.}, \bibinfo{author}{Zikic, D.},
  \bibinfo{author}{Ghosh, A.}, \bibinfo{author}{Tanno, R.},
  \bibinfo{author}{Wottschel, V.}, \bibinfo{author}{Zhang, J.},
  \bibinfo{author}{Kaden, E.}, \bibinfo{author}{Dyrby, T.B.},
  \bibinfo{author}{Sotiropoulos, S.N.}, \bibinfo{author}{Zhang, H.},
  \bibinfo{author}{Criminisi, A.}, \bibinfo{year}{2017}.
\newblock \bibinfo{title}{Image quality transfer and applications in diffusion
  {MRI}}.
\newblock \bibinfo{journal}{NeuroImage} \bibinfo{volume}{152},
  \bibinfo{pages}{283--298}.
\newblock \URLprefix
  \url{http://www.sciencedirect.com/science/article/pii/S1053811917302008},
  \DOIprefix\doi{10.1016/j.neuroimage.2017.02.089}.
\bibitem[{Andersson and Sotiropoulos(2016)}]{andersson_integrated_2016}
\bibinfo{author}{Andersson, J.L.R.}, \bibinfo{author}{Sotiropoulos, S.N.},
  \bibinfo{year}{2016}.
\newblock \bibinfo{title}{An integrated approach to correction for
  off-resonance effects and subject movement in diffusion {MR} imaging}.
\newblock \bibinfo{journal}{NeuroImage} \bibinfo{volume}{125},
  \bibinfo{pages}{1063--1078}.
\newblock \URLprefix
  \url{http://www.sciencedirect.com/science/article/pii/S1053811915009209},
  \DOIprefix\doi{10.1016/j.neuroimage.2015.10.019}.
\bibitem[{Ankele et~al.(2017)Ankele, Lim, Groeschel and
  Schultz}]{ankele_versatile_2017}
\bibinfo{author}{Ankele, M.}, \bibinfo{author}{Lim, L.H.},
  \bibinfo{author}{Groeschel, S.}, \bibinfo{author}{Schultz, T.},
  \bibinfo{year}{2017}.
\newblock \bibinfo{title}{Versatile, robust, and efficient tractography with
  constrained higher-order tensor {fODFs}}.
\newblock \bibinfo{journal}{International Journal of Computer Assisted
  Radiology and Surgery} \bibinfo{volume}{12}, \bibinfo{pages}{1257--1270}.
\newblock \DOIprefix\doi{10.1007/s11548-017-1593-6}.
\bibitem[{Avants et~al.(2008)Avants, Epstein, Grossman and
  Gee}]{avants_symmetric_2008}
\bibinfo{author}{Avants, B.B.}, \bibinfo{author}{Epstein, C.L.},
  \bibinfo{author}{Grossman, M.}, \bibinfo{author}{Gee, J.C.},
  \bibinfo{year}{2008}.
\newblock \bibinfo{title}{Symmetric diffeomorphic image registration with
  cross-correlation: evaluating automated labeling of elderly and
  neurodegenerative brain}.
\newblock \bibinfo{journal}{Medical Image Analysis} \bibinfo{volume}{12},
  \bibinfo{pages}{26--41}.
\newblock \DOIprefix\doi{10.1016/j.media.2007.06.004}.
\bibitem[{Basser et~al.(2000)Basser, Pajevic, Pierpaoli, Duda and
  Aldroubi}]{basser_vivo_2000}
\bibinfo{author}{Basser, P.J.}, \bibinfo{author}{Pajevic, S.},
  \bibinfo{author}{Pierpaoli, C.}, \bibinfo{author}{Duda, J.},
  \bibinfo{author}{Aldroubi, A.}, \bibinfo{year}{2000}.
\newblock \bibinfo{title}{In vivo fiber tractography using {DT}-{MRI} data}.
\newblock \bibinfo{journal}{Magnetic Resonance in Medicine}
  \bibinfo{volume}{44}, \bibinfo{pages}{625--632}.
\bibitem[{Behrens et~al.(2003)Behrens, Woolrich, Jenkinson, Johansen-Berg,
  Nunes, Clare, Matthews, Brady and Smith}]{behrens_characterization_2003}
\bibinfo{author}{Behrens, T.E.J.}, \bibinfo{author}{Woolrich, M.W.},
  \bibinfo{author}{Jenkinson, M.}, \bibinfo{author}{Johansen-Berg, H.},
  \bibinfo{author}{Nunes, R.G.}, \bibinfo{author}{Clare, S.},
  \bibinfo{author}{Matthews, P.M.}, \bibinfo{author}{Brady, J.M.},
  \bibinfo{author}{Smith, S.M.}, \bibinfo{year}{2003}.
\newblock \bibinfo{title}{Characterization and propagation of uncertainty in
  diffusion-weighted {MR} imaging}.
\newblock \bibinfo{journal}{Magnetic Resonance in Medicine}
  \bibinfo{volume}{50}, \bibinfo{pages}{1077--1088}.
\newblock \DOIprefix\doi{10.1002/mrm.10609}.
\bibitem[{Bustillo et~al.(2017)Bustillo, Jones, Chen, Lemke, Abbott, Qualls,
  Stromberg, Canive and Gasparovic}]{bustillo_glutamatergic_2017}
\bibinfo{author}{Bustillo, J.R.}, \bibinfo{author}{Jones, T.},
  \bibinfo{author}{Chen, H.}, \bibinfo{author}{Lemke, N.},
  \bibinfo{author}{Abbott, C.}, \bibinfo{author}{Qualls, C.},
  \bibinfo{author}{Stromberg, S.}, \bibinfo{author}{Canive, J.},
  \bibinfo{author}{Gasparovic, C.}, \bibinfo{year}{2017}.
\newblock \bibinfo{title}{Glutamatergic and {Neuronal} {Dysfunction} in {Gray}
  and {White} {Matter}: {A} {Spectroscopic} {Imaging} {Study} in a {Large}
  {Schizophrenia} {Sample}}.
\newblock \bibinfo{journal}{Schizophrenia Bulletin} \bibinfo{volume}{43},
  \bibinfo{pages}{611--619}.
\newblock \DOIprefix\doi{10.1093/schbul/sbw122}.
\bibitem[{Desikan et~al.(2006)Desikan, Ségonne, Fischl, Quinn, Dickerson,
  Blacker, Buckner, Dale, Maguire, Hyman, Albert and
  Killiany}]{desikan_automated_2006}
\bibinfo{author}{Desikan, R.S.}, \bibinfo{author}{Ségonne, F.},
  \bibinfo{author}{Fischl, B.}, \bibinfo{author}{Quinn, B.T.},
  \bibinfo{author}{Dickerson, B.C.}, \bibinfo{author}{Blacker, D.},
  \bibinfo{author}{Buckner, R.L.}, \bibinfo{author}{Dale, A.M.},
  \bibinfo{author}{Maguire, R.P.}, \bibinfo{author}{Hyman, B.T.},
  \bibinfo{author}{Albert, M.S.}, \bibinfo{author}{Killiany, R.J.},
  \bibinfo{year}{2006}.
\newblock \bibinfo{title}{An automated labeling system for subdividing the
  human cerebral cortex on {MRI} scans into gyral based regions of interest}.
\newblock \bibinfo{journal}{NeuroImage} \bibinfo{volume}{31},
  \bibinfo{pages}{968--980}.
\newblock \DOIprefix\doi{10.1016/j.neuroimage.2006.01.021}.
\bibitem[{Dumoulin and Visin(2016)}]{dumoulin_guide_2016}
\bibinfo{author}{Dumoulin, V.}, \bibinfo{author}{Visin, F.},
  \bibinfo{year}{2016}.
\newblock \bibinfo{title}{A guide to convolution arithmetic for deep learning}.
\newblock \bibinfo{journal}{arXiv:1603.07285 [cs, stat]} \URLprefix
  \url{http://arxiv.org/abs/1603.07285}. \bibinfo{note}{arXiv: 1603.07285}.
\bibitem[{Ester et~al.(1996)Ester, Kriegel, Sander and
  Xu}]{ester_density-based_1996}
\bibinfo{author}{Ester, M.}, \bibinfo{author}{Kriegel, H.P.},
  \bibinfo{author}{Sander, J.}, \bibinfo{author}{Xu, X.}, \bibinfo{year}{1996}.
\newblock \bibinfo{title}{A density-based algorithm for discovering clusters in
  large spatial databases with noise.}, in: \bibinfo{booktitle}{Kdd}, pp.
  \bibinfo{pages}{226--231}.
\bibitem[{Garyfallidis et~al.(2014)Garyfallidis, Brett, Amirbekian, Rokem, Van
  Der~Walt, Descoteaux and Nimmo-Smith}]{garyfallidis_dipy_2014}
\bibinfo{author}{Garyfallidis, E.}, \bibinfo{author}{Brett, M.},
  \bibinfo{author}{Amirbekian, B.}, \bibinfo{author}{Rokem, A.},
  \bibinfo{author}{Van Der~Walt, S.}, \bibinfo{author}{Descoteaux, M.},
  \bibinfo{author}{Nimmo-Smith, I.}, \bibinfo{year}{2014}.
\newblock \bibinfo{title}{Dipy, a library for the analysis of diffusion {MRI}
  data}.
\newblock \bibinfo{journal}{Frontiers in Neuroinformatics} \bibinfo{volume}{8}.
\newblock \URLprefix
  \url{https://www.frontiersin.org/articles/10.3389/fninf.2014.00008/full},
  \DOIprefix\doi{10.3389/fninf.2014.00008}.
\bibitem[{Garyfallidis et~al.(2017)Garyfallidis, Côté, Rheault, Sidhu, Hau,
  Petit, Fortin, Cunanne and Descoteaux}]{garyfallidis_recognition_2017}
\bibinfo{author}{Garyfallidis, E.}, \bibinfo{author}{Côté, M.A.},
  \bibinfo{author}{Rheault, F.}, \bibinfo{author}{Sidhu, J.},
  \bibinfo{author}{Hau, J.}, \bibinfo{author}{Petit, L.},
  \bibinfo{author}{Fortin, D.}, \bibinfo{author}{Cunanne, S.},
  \bibinfo{author}{Descoteaux, M.}, \bibinfo{year}{2017}.
\newblock \bibinfo{title}{Recognition of white matter bundles using local and
  global streamline-based registration and clustering}.
\newblock \bibinfo{journal}{NeuroImage}
  \DOIprefix\doi{10.1016/j.neuroimage.2017.07.015}.
\bibitem[{Hirjak et~al.(2017)Hirjak, Thomann, Wolf, Kubera, Goch, Hering and
  Maier-Hein}]{hirjak_white_2017}
\bibinfo{author}{Hirjak, D.}, \bibinfo{author}{Thomann, P.A.},
  \bibinfo{author}{Wolf, R.C.}, \bibinfo{author}{Kubera, K.M.},
  \bibinfo{author}{Goch, C.}, \bibinfo{author}{Hering, J.},
  \bibinfo{author}{Maier-Hein, K.H.}, \bibinfo{year}{2017}.
\newblock \bibinfo{title}{White matter microstructure variations contribute to
  neurological soft signs in healthy adults}.
\newblock \bibinfo{journal}{Human Brain Mapping}
  \DOIprefix\doi{10.1002/hbm.23609}.
\bibitem[{Isensee et~al.(2018)Isensee, Kickingereder, Wick, Bendszus and
  Maier-Hein}]{Isensee18}
\bibinfo{author}{Isensee, F.}, \bibinfo{author}{Kickingereder, P.},
  \bibinfo{author}{Wick, W.}, \bibinfo{author}{Bendszus, M.},
  \bibinfo{author}{Maier-Hein, K.H.}, \bibinfo{year}{2018}.
\newblock \bibinfo{title}{Brain {Tumor} {Segmentation} and {Radiomics}
  {Survival} {Prediction}: {Contribution} to the {BRATS} 2017 {Challenge}}.
\newblock \bibinfo{journal}{arXiv:1802.10508 [cs]} \URLprefix
  \url{http://arxiv.org/abs/1802.10508}. \bibinfo{note}{arXiv: 1802.10508}.
\bibitem[{Jenkinson et~al.(2012)Jenkinson, Beckmann, Behrens, Woolrich and
  Smith}]{jenkinson_fsl_2012}
\bibinfo{author}{Jenkinson, M.}, \bibinfo{author}{Beckmann, C.F.},
  \bibinfo{author}{Behrens, T.E.J.}, \bibinfo{author}{Woolrich, M.W.},
  \bibinfo{author}{Smith, S.M.}, \bibinfo{year}{2012}.
\newblock \bibinfo{title}{{FSL}}.
\newblock \bibinfo{journal}{NeuroImage} \bibinfo{volume}{62},
  \bibinfo{pages}{782--790}.
\newblock \URLprefix
  \url{http://www.sciencedirect.com/science/article/pii/S1053811911010603},
  \DOIprefix\doi{10.1016/j.neuroimage.2011.09.015}.
\bibitem[{Jeurissen et~al.(2014)Jeurissen, Tournier, Dhollander, Connelly and
  Sijbers}]{jeurissen_multi-tissue_2014}
\bibinfo{author}{Jeurissen, B.}, \bibinfo{author}{Tournier, J.D.},
  \bibinfo{author}{Dhollander, T.}, \bibinfo{author}{Connelly, A.},
  \bibinfo{author}{Sijbers, J.}, \bibinfo{year}{2014}.
\newblock \bibinfo{title}{Multi-tissue constrained spherical deconvolution for
  improved analysis of multi-shell diffusion {MRI} data}.
\newblock \bibinfo{journal}{NeuroImage} \bibinfo{volume}{103},
  \bibinfo{pages}{411--426}.
\newblock \DOIprefix\doi{10.1016/j.neuroimage.2014.07.061}.
\bibitem[{Jones et~al.(2001)Jones, Oliphant, Peterson
  et~al.}]{jones_scipy_2001}
\bibinfo{author}{Jones, E.}, \bibinfo{author}{Oliphant, T.},
  \bibinfo{author}{Peterson, P.}, et~al., \bibinfo{year}{2001}.
\newblock \bibinfo{title}{{SciPy}: Open source scientific tools for {Python}}.
\newblock \URLprefix \url{http://www.scipy.org/}. \bibinfo{note}{[Online;
  accessed <today>]}.
\bibitem[{Kingma and Ba(2014)}]{kingma_adam:_2014}
\bibinfo{author}{Kingma, D.P.}, \bibinfo{author}{Ba, J.}, \bibinfo{year}{2014}.
\newblock \bibinfo{title}{Adam: {A} {Method} for {Stochastic} {Optimization}}.
\newblock \bibinfo{journal}{arXiv:1412.6980 [cs]} \URLprefix
  \url{http://arxiv.org/abs/1412.6980}. \bibinfo{note}{arXiv: 1412.6980}.
\bibitem[{Knösche et~al.(2015)Knösche, Anwander, Liptrot and
  Dyrby}]{knosche_validation_2015}
\bibinfo{author}{Knösche, T.R.}, \bibinfo{author}{Anwander, A.},
  \bibinfo{author}{Liptrot, M.}, \bibinfo{author}{Dyrby, T.B.},
  \bibinfo{year}{2015}.
\newblock \bibinfo{title}{Validation of tractography: {Comparison} with
  manganese tracing}.
\newblock \bibinfo{journal}{Human Brain Mapping} \bibinfo{volume}{36},
  \bibinfo{pages}{4116--4134}.
\newblock \DOIprefix\doi{10.1002/hbm.22902}.
\bibitem[{Maier-Hein et~al.(2017)Maier-Hein, Neher, Houde, Côté,
  Garyfallidis, Zhong, Chamberland, Yeh, Lin, Ji, Reddick, Glass, Chen, Feng,
  Gao, Wu, Ma, Renjie, Li, Westin, Deslauriers-Gauthier, González, Paquette,
  St-Jean, Girard, Rheault, Sidhu, Tax, Guo, Mesri, Dávid, Froeling,
  Heemskerk, Leemans, Boré, Pinsard, Bedetti, Desrosiers, Brambati, Doyon,
  Sarica, Vasta, Cerasa, Quattrone, Yeatman, Khan, Hodges, Alexander,
  Romascano, Barakovic, Auría, Esteban, Lemkaddem, Thiran, Cetingul, Odry,
  Mailhe, Nadar, Pizzagalli, Prasad, Villalon-Reina, Galvis, Thompson, Requejo,
  Laguna, Lacerda, Barrett, Dell’Acqua, Catani, Petit, Caruyer, Daducci,
  Dyrby, Holland-Letz, Hilgetag, Stieltjes and
  Descoteaux}]{maier-hein_challenge_2017}
\bibinfo{author}{Maier-Hein, K.H.}, \bibinfo{author}{Neher, P.F.},
  \bibinfo{author}{Houde, J.C.}, \bibinfo{author}{Côté, M.A.},
  \bibinfo{author}{Garyfallidis, E.}, \bibinfo{author}{Zhong, J.},
  \bibinfo{author}{Chamberland, M.}, \bibinfo{author}{Yeh, F.C.},
  \bibinfo{author}{Lin, Y.C.}, \bibinfo{author}{Ji, Q.},
  \bibinfo{author}{Reddick, W.E.}, \bibinfo{author}{Glass, J.O.},
  \bibinfo{author}{Chen, D.Q.}, \bibinfo{author}{Feng, Y.},
  \bibinfo{author}{Gao, C.}, \bibinfo{author}{Wu, Y.}, \bibinfo{author}{Ma,
  J.}, \bibinfo{author}{Renjie, H.}, \bibinfo{author}{Li, Q.},
  \bibinfo{author}{Westin, C.F.}, \bibinfo{author}{Deslauriers-Gauthier, S.},
  \bibinfo{author}{González, J.O.O.}, \bibinfo{author}{Paquette, M.},
  \bibinfo{author}{St-Jean, S.}, \bibinfo{author}{Girard, G.},
  \bibinfo{author}{Rheault, F.}, \bibinfo{author}{Sidhu, J.},
  \bibinfo{author}{Tax, C.M.W.}, \bibinfo{author}{Guo, F.},
  \bibinfo{author}{Mesri, H.Y.}, \bibinfo{author}{Dávid, S.},
  \bibinfo{author}{Froeling, M.}, \bibinfo{author}{Heemskerk, A.M.},
  \bibinfo{author}{Leemans, A.}, \bibinfo{author}{Boré, A.},
  \bibinfo{author}{Pinsard, B.}, \bibinfo{author}{Bedetti, C.},
  \bibinfo{author}{Desrosiers, M.}, \bibinfo{author}{Brambati, S.},
  \bibinfo{author}{Doyon, J.}, \bibinfo{author}{Sarica, A.},
  \bibinfo{author}{Vasta, R.}, \bibinfo{author}{Cerasa, A.},
  \bibinfo{author}{Quattrone, A.}, \bibinfo{author}{Yeatman, J.},
  \bibinfo{author}{Khan, A.R.}, \bibinfo{author}{Hodges, W.},
  \bibinfo{author}{Alexander, S.}, \bibinfo{author}{Romascano, D.},
  \bibinfo{author}{Barakovic, M.}, \bibinfo{author}{Auría, A.},
  \bibinfo{author}{Esteban, O.}, \bibinfo{author}{Lemkaddem, A.},
  \bibinfo{author}{Thiran, J.P.}, \bibinfo{author}{Cetingul, H.E.},
  \bibinfo{author}{Odry, B.L.}, \bibinfo{author}{Mailhe, B.},
  \bibinfo{author}{Nadar, M.S.}, \bibinfo{author}{Pizzagalli, F.},
  \bibinfo{author}{Prasad, G.}, \bibinfo{author}{Villalon-Reina, J.E.},
  \bibinfo{author}{Galvis, J.}, \bibinfo{author}{Thompson, P.M.},
  \bibinfo{author}{Requejo, F.D.S.}, \bibinfo{author}{Laguna, P.L.},
  \bibinfo{author}{Lacerda, L.M.}, \bibinfo{author}{Barrett, R.},
  \bibinfo{author}{Dell’Acqua, F.}, \bibinfo{author}{Catani, M.},
  \bibinfo{author}{Petit, L.}, \bibinfo{author}{Caruyer, E.},
  \bibinfo{author}{Daducci, A.}, \bibinfo{author}{Dyrby, T.B.},
  \bibinfo{author}{Holland-Letz, T.}, \bibinfo{author}{Hilgetag, C.C.},
  \bibinfo{author}{Stieltjes, B.}, \bibinfo{author}{Descoteaux, M.},
  \bibinfo{year}{2017}.
\newblock \bibinfo{title}{The challenge of mapping the human connectome based
  on diffusion tractography}.
\newblock \bibinfo{journal}{Nature Communications} \bibinfo{volume}{8},
  \bibinfo{pages}{1349}.
\newblock \URLprefix \url{https://www.nature.com/articles/s41467-017-01285-x},
  \DOIprefix\doi{10.1038/s41467-017-01285-x}.
\bibitem[{Mori et~al.(1999)Mori, Crain, Chacko and van
  Zijl}]{mori_three-dimensional_1999}
\bibinfo{author}{Mori, S.}, \bibinfo{author}{Crain, B.J.},
  \bibinfo{author}{Chacko, V.P.}, \bibinfo{author}{van Zijl, P.C.},
  \bibinfo{year}{1999}.
\newblock \bibinfo{title}{Three-dimensional tracking of axonal projections in
  the brain by magnetic resonance imaging}.
\newblock \bibinfo{journal}{Annals of Neurology} \bibinfo{volume}{45},
  \bibinfo{pages}{265--269}.
\bibitem[{Nair and Hinton(2010)}]{nair_rectified_2010}
\bibinfo{author}{Nair, V.}, \bibinfo{author}{Hinton, G.}, \bibinfo{year}{2010}.
\newblock \bibinfo{title}{Rectified {Linear} {Units} {Improve} {Restricted}
  {Boltzmann} {Machines}}.
\newblock \bibinfo{journal}{ICML} .
\bibitem[{Neher et~al.(2014)Neher, Laun, Stieltjes and
  Maier-Hein}]{neher_fiberfox:_2014}
\bibinfo{author}{Neher, P.F.}, \bibinfo{author}{Laun, F.B.},
  \bibinfo{author}{Stieltjes, B.}, \bibinfo{author}{Maier-Hein, K.H.},
  \bibinfo{year}{2014}.
\newblock \bibinfo{title}{Fiberfox: facilitating the creation of realistic
  white matter software phantoms}.
\newblock \bibinfo{journal}{Magnetic resonance in medicine}
  \bibinfo{volume}{72}, \bibinfo{pages}{1460--1470}.
\bibitem[{Nooner et~al.(2012)Nooner, Colcombe, Tobe, Mennes, Benedict, Moreno,
  Panek, Brown, Zavitz, Li, Sikka, Gutman, Bangaru, Schlachter, Kamiel, Anwar,
  Hinz, Kaplan, Rachlin, Adelsberg, Cheung, Khanuja, Yan, Craddock, Calhoun,
  Courtney, King, Wood, Cox, Kelly, Di~Martino, Petkova, Reiss, Duan, Thomsen,
  Biswal, Coffey, Hoptman, Javitt, Pomara, Sidtis, Koplewicz, Castellanos,
  Leventhal and Milham}]{nooner_nki-rockland_2012}
\bibinfo{author}{Nooner, K.B.}, \bibinfo{author}{Colcombe, S.J.},
  \bibinfo{author}{Tobe, R.H.}, \bibinfo{author}{Mennes, M.},
  \bibinfo{author}{Benedict, M.M.}, \bibinfo{author}{Moreno, A.L.},
  \bibinfo{author}{Panek, L.J.}, \bibinfo{author}{Brown, S.},
  \bibinfo{author}{Zavitz, S.T.}, \bibinfo{author}{Li, Q.},
  \bibinfo{author}{Sikka, S.}, \bibinfo{author}{Gutman, D.},
  \bibinfo{author}{Bangaru, S.}, \bibinfo{author}{Schlachter, R.T.},
  \bibinfo{author}{Kamiel, S.M.}, \bibinfo{author}{Anwar, A.R.},
  \bibinfo{author}{Hinz, C.M.}, \bibinfo{author}{Kaplan, M.S.},
  \bibinfo{author}{Rachlin, A.B.}, \bibinfo{author}{Adelsberg, S.},
  \bibinfo{author}{Cheung, B.}, \bibinfo{author}{Khanuja, R.},
  \bibinfo{author}{Yan, C.}, \bibinfo{author}{Craddock, C.C.},
  \bibinfo{author}{Calhoun, V.}, \bibinfo{author}{Courtney, W.},
  \bibinfo{author}{King, M.}, \bibinfo{author}{Wood, D.}, \bibinfo{author}{Cox,
  C.L.}, \bibinfo{author}{Kelly, A.M.C.}, \bibinfo{author}{Di~Martino, A.},
  \bibinfo{author}{Petkova, E.}, \bibinfo{author}{Reiss, P.T.},
  \bibinfo{author}{Duan, N.}, \bibinfo{author}{Thomsen, D.},
  \bibinfo{author}{Biswal, B.}, \bibinfo{author}{Coffey, B.},
  \bibinfo{author}{Hoptman, M.J.}, \bibinfo{author}{Javitt, D.C.},
  \bibinfo{author}{Pomara, N.}, \bibinfo{author}{Sidtis, J.J.},
  \bibinfo{author}{Koplewicz, H.S.}, \bibinfo{author}{Castellanos, F.X.},
  \bibinfo{author}{Leventhal, B.L.}, \bibinfo{author}{Milham, M.P.},
  \bibinfo{year}{2012}.
\newblock \bibinfo{title}{The {NKI}-{Rockland} {Sample}: {A} {Model} for
  {Accelerating} the {Pace} of {Discovery} {Science} in {Psychiatry}}.
\newblock \bibinfo{journal}{Frontiers in Neuroscience} \bibinfo{volume}{6},
  \bibinfo{pages}{152}.
\newblock \DOIprefix\doi{10.3389/fnins.2012.00152}.
\bibitem[{O'Donnell and Westin(2007)}]{odonnell_automatic_2007}
\bibinfo{author}{O'Donnell, L.J.}, \bibinfo{author}{Westin, C.F.},
  \bibinfo{year}{2007}.
\newblock \bibinfo{title}{Automatic tractography segmentation using a
  high-dimensional white matter atlas}.
\newblock \bibinfo{journal}{IEEE transactions on medical imaging}
  \bibinfo{volume}{26}, \bibinfo{pages}{1562--1575}.
\newblock \DOIprefix\doi{10.1109/TMI.2007.906785}.
\bibitem[{O’Donnell et~al.(2016)O’Donnell, Suter, Rigolo, Kahali, Zhang,
  Norton, Albi, Olubiyi, Meola, Essayed, Unadkat, Ciris, Wells, Rathi, Westin
  and Golby}]{odonnell_automated_2016}
\bibinfo{author}{O’Donnell, L.J.}, \bibinfo{author}{Suter, Y.},
  \bibinfo{author}{Rigolo, L.}, \bibinfo{author}{Kahali, P.},
  \bibinfo{author}{Zhang, F.}, \bibinfo{author}{Norton, I.},
  \bibinfo{author}{Albi, A.}, \bibinfo{author}{Olubiyi, O.},
  \bibinfo{author}{Meola, A.}, \bibinfo{author}{Essayed, W.I.},
  \bibinfo{author}{Unadkat, P.}, \bibinfo{author}{Ciris, P.A.},
  \bibinfo{author}{Wells, W.M.}, \bibinfo{author}{Rathi, Y.},
  \bibinfo{author}{Westin, C.F.}, \bibinfo{author}{Golby, A.J.},
  \bibinfo{year}{2016}.
\newblock \bibinfo{title}{Automated white matter fiber tract identification in
  patients with brain tumors}.
\newblock \bibinfo{journal}{NeuroImage : Clinical} \bibinfo{volume}{13},
  \bibinfo{pages}{138--153}.
\newblock \URLprefix
  \url{https://www.ncbi.nlm.nih.gov/pmc/articles/PMC5144756/},
  \DOIprefix\doi{10.1016/j.nicl.2016.11.023}.
\bibitem[{Poldrack et~al.(2016)Poldrack, Congdon, Triplett, Gorgolewski,
  Karlsgodt, Mumford, Sabb, Freimer, London, Cannon and
  Bilder}]{poldrack_phenome-wide_2016}
\bibinfo{author}{Poldrack, R.A.}, \bibinfo{author}{Congdon, E.},
  \bibinfo{author}{Triplett, W.}, \bibinfo{author}{Gorgolewski, K.J.},
  \bibinfo{author}{Karlsgodt, K.H.}, \bibinfo{author}{Mumford, J.A.},
  \bibinfo{author}{Sabb, F.W.}, \bibinfo{author}{Freimer, N.B.},
  \bibinfo{author}{London, E.D.}, \bibinfo{author}{Cannon, T.D.},
  \bibinfo{author}{Bilder, R.M.}, \bibinfo{year}{2016}.
\newblock \bibinfo{title}{A phenome-wide examination of neural and cognitive
  function}.
\newblock \bibinfo{journal}{Scientific Data} \bibinfo{volume}{3},
  \bibinfo{pages}{160110}.
\newblock \URLprefix \url{https://www.nature.com/articles/sdata2016110},
  \DOIprefix\doi{10.1038/sdata.2016.110}.
\bibitem[{Raffelt et~al.(2011)Raffelt, Tournier, Fripp, Crozier, Connelly and
  Salvado}]{raffelt_symmetric_2011}
\bibinfo{author}{Raffelt, D.}, \bibinfo{author}{Tournier, J.D.},
  \bibinfo{author}{Fripp, J.}, \bibinfo{author}{Crozier, S.},
  \bibinfo{author}{Connelly, A.}, \bibinfo{author}{Salvado, O.},
  \bibinfo{year}{2011}.
\newblock \bibinfo{title}{Symmetric diffeomorphic registration of fibre
  orientation distributions}.
\newblock \bibinfo{journal}{NeuroImage} \bibinfo{volume}{56},
  \bibinfo{pages}{1171--1180}.
\newblock \URLprefix
  \url{http://www.sciencedirect.com/science/article/pii/S1053811911001534},
  \DOIprefix\doi{10.1016/j.neuroimage.2011.02.014}.
\bibitem[{Rheault et~al.(2018)Rheault, St-Onge, Sidhu, Maier-Hein,
  Tzourio-Mazoyer, Petit and Descoteaux}]{rheault_bundle-specific_2018}
\bibinfo{author}{Rheault, F.}, \bibinfo{author}{St-Onge, E.},
  \bibinfo{author}{Sidhu, J.}, \bibinfo{author}{Maier-Hein, K.},
  \bibinfo{author}{Tzourio-Mazoyer, N.}, \bibinfo{author}{Petit, L.},
  \bibinfo{author}{Descoteaux, M.}, \bibinfo{year}{2018}.
\newblock \bibinfo{title}{Bundle-specific tractography with incorporated
  anatomical and orientational priors}.
\newblock \bibinfo{journal}{NeuroImage} \bibinfo{volume}{186},
  \bibinfo{pages}{382--398}.
\newblock \DOIprefix\doi{10.1016/j.neuroimage.2018.11.018}.
\bibitem[{Rokem et~al.(2013)Rokem, Yeatman, Pestilli and
  Wandell}]{rokem_high_2013}
\bibinfo{author}{Rokem, A.}, \bibinfo{author}{Yeatman, J.},
  \bibinfo{author}{Pestilli, F.}, \bibinfo{author}{Wandell, B.},
  \bibinfo{year}{2013}.
\newblock \bibinfo{title}{High angular resolution diffusion {MRI}}.
\newblock \URLprefix \url{https://purl.stanford.edu/yx282xq2090}.
\bibitem[{Ronneberger et~al.(2015)Ronneberger, Fischer and
  Brox}]{ronneberger_u-net_2015}
\bibinfo{author}{Ronneberger, O.}, \bibinfo{author}{Fischer, P.},
  \bibinfo{author}{Brox, T.}, \bibinfo{year}{2015}.
\newblock \bibinfo{title}{U-{Net}: {Convolutional} {Networks} for {Biomedical}
  {Image} {Segmentation}}.
\newblock \bibinfo{journal}{arXiv:1505.04597 [cs]} \URLprefix
  \url{http://arxiv.org/abs/1505.04597}. \bibinfo{note}{arXiv: 1505.04597}.
\bibitem[{Stieltjes et~al.(2013)Stieltjes, Brunner, Maier-Hein and
  Laun}]{stieltjes_diffusion_2013}
\bibinfo{author}{Stieltjes, B.}, \bibinfo{author}{Brunner, R.},
  \bibinfo{author}{Maier-Hein, K.}, \bibinfo{author}{Laun, F.},
  \bibinfo{year}{2013}.
\newblock \bibinfo{title}{Diffusion tensor imaging: introduction and atlas}.
\bibitem[{Taha and Hanbury(2015)}]{taha_metrics_2015}
\bibinfo{author}{Taha, A.A.}, \bibinfo{author}{Hanbury, A.},
  \bibinfo{year}{2015}.
\newblock \bibinfo{title}{Metrics for evaluating 3d medical image segmentation:
  analysis, selection, and tool}.
\newblock \bibinfo{journal}{BMC Medical Imaging} \bibinfo{volume}{15}.
\newblock \URLprefix
  \url{https://www.ncbi.nlm.nih.gov/pmc/articles/PMC4533825/},
  \DOIprefix\doi{10.1186/s12880-015-0068-x}.
\bibitem[{Tisdall et~al.(2012)Tisdall, Hess, Reuter, Meintjes, Fischl and
  van~der Kouwe}]{tisdall_volumetric_2012}
\bibinfo{author}{Tisdall, M.D.}, \bibinfo{author}{Hess, A.T.},
  \bibinfo{author}{Reuter, M.}, \bibinfo{author}{Meintjes, E.M.},
  \bibinfo{author}{Fischl, B.}, \bibinfo{author}{van~der Kouwe, A.J.W.},
  \bibinfo{year}{2012}.
\newblock \bibinfo{title}{Volumetric navigators for prospective motion
  correction and selective reacquisition in neuroanatomical {MRI}}.
\newblock \bibinfo{journal}{Magnetic Resonance in Medicine}
  \bibinfo{volume}{68}, \bibinfo{pages}{389--399}.
\newblock \DOIprefix\doi{10.1002/mrm.23228}.
\bibitem[{Tournier et~al.(2007)Tournier, Calamante and
  Connelly}]{tournier_robust_2007}
\bibinfo{author}{Tournier, J.D.}, \bibinfo{author}{Calamante, F.},
  \bibinfo{author}{Connelly, A.}, \bibinfo{year}{2007}.
\newblock \bibinfo{title}{Robust determination of the fibre orientation
  distribution in diffusion {MRI}: non-negativity constrained super-resolved
  spherical deconvolution}.
\newblock \bibinfo{journal}{NeuroImage} \bibinfo{volume}{35},
  \bibinfo{pages}{1459--1472}.
\newblock \DOIprefix\doi{10.1016/j.neuroimage.2007.02.016}.
\bibitem[{Van~Essen et~al.(2013)Van~Essen, Smith, Barch, Behrens, Yacoub and
  Ugurbil}]{van_essen_wu-minn_2013}
\bibinfo{author}{Van~Essen, D.C.}, \bibinfo{author}{Smith, S.M.},
  \bibinfo{author}{Barch, D.M.}, \bibinfo{author}{Behrens, T.E.J.},
  \bibinfo{author}{Yacoub, E.}, \bibinfo{author}{Ugurbil, K.},
  \bibinfo{year}{2013}.
\newblock \bibinfo{title}{The {WU}-{Minn} {Human} {Connectome} {Project}: {An}
  overview}.
\newblock \bibinfo{journal}{NeuroImage} \bibinfo{volume}{80},
  \bibinfo{pages}{62--79}.
\newblock \URLprefix
  \url{http://www.sciencedirect.com/science/article/pii/S1053811913005351},
  \DOIprefix\doi{10.1016/j.neuroimage.2013.05.041}.
\bibitem[{Veraart et~al.(2016)Veraart, Novikov, Christiaens, Ades-aron, Sijbers
  and Fieremans}]{veraart_denoising_2016}
\bibinfo{author}{Veraart, J.}, \bibinfo{author}{Novikov, D.S.},
  \bibinfo{author}{Christiaens, D.}, \bibinfo{author}{Ades-aron, B.},
  \bibinfo{author}{Sijbers, J.}, \bibinfo{author}{Fieremans, E.},
  \bibinfo{year}{2016}.
\newblock \bibinfo{title}{Denoising of diffusion {MRI} using random matrix
  theory}.
\newblock \bibinfo{journal}{NeuroImage} \bibinfo{volume}{142},
  \bibinfo{pages}{394--406}.
\newblock \URLprefix
  \url{http://www.sciencedirect.com/science/article/pii/S1053811916303949},
  \DOIprefix\doi{10.1016/j.neuroimage.2016.08.016}.
\bibitem[{Wassermann et~al.(2016)Wassermann, Makris, Rathi, Shenton, Kikinis,
  Kubicki and Westin}]{wassermann_white_2016}
\bibinfo{author}{Wassermann, D.}, \bibinfo{author}{Makris, N.},
  \bibinfo{author}{Rathi, Y.}, \bibinfo{author}{Shenton, M.},
  \bibinfo{author}{Kikinis, R.}, \bibinfo{author}{Kubicki, M.},
  \bibinfo{author}{Westin, C.F.}, \bibinfo{year}{2016}.
\newblock \bibinfo{title}{The white matter query language: a novel approach for
  describing human white matter anatomy}.
\newblock \bibinfo{journal}{Brain Structure and Function}
  \bibinfo{volume}{221}, \bibinfo{pages}{4705--4721}.
\newblock \URLprefix
  \url{https://link.springer.com/article/10.1007/s00429-015-1179-4}.
\bibitem[{Wasserthal et~al.(2018a)Wasserthal, Neher and
  Maier-Hein}]{Wasserthal18b}
\bibinfo{author}{Wasserthal, J.}, \bibinfo{author}{Neher, P.},
  \bibinfo{author}{Maier-Hein, K.}, \bibinfo{year}{2018}a.
\newblock \bibinfo{title}{Tract {Orientation} {Mapping} for {Bundle}-{Specific}
  {Tractography}: 21st {International Conference on Medical Image Computing and
  Computer Assisted Intervention (MICCAI)}, {Granada}, {Spain}, {September}
  16-20, 2018, {Proceedings}, {Part} {III}}, pp. \bibinfo{pages}{36--44}.
\newblock \DOIprefix\doi{10.1007/978-3-030-00931-1_5}.
\bibitem[{Wasserthal et~al.(2018b)Wasserthal, Neher and
  Maier-Hein}]{Wasserthal18a}
\bibinfo{author}{Wasserthal, J.}, \bibinfo{author}{Neher, P.},
  \bibinfo{author}{Maier-Hein, K.H.}, \bibinfo{year}{2018}b.
\newblock \bibinfo{title}{{TractSeg} - {Fast} and accurate white matter tract
  segmentation}.
\newblock \bibinfo{journal}{NeuroImage}
  \DOIprefix\doi{10.1016/j.neuroimage.2018.07.070}.
\bibitem[{Wilcoxon(1945)}]{wilcoxon_individual_1945}
\bibinfo{author}{Wilcoxon, F.}, \bibinfo{year}{1945}.
\newblock \bibinfo{title}{Individual {Comparisons} by {Ranking} {Methods}}.
\newblock \bibinfo{journal}{Biometrics Bulletin} \bibinfo{volume}{1},
  \bibinfo{pages}{80--83}.
\newblock \URLprefix \url{http://www.jstor.org/stable/3001968},
  \DOIprefix\doi{10.2307/3001968}.
\bibitem[{Yendiki et~al.(2011)Yendiki, Panneck, Srinivasan, Stevens, Zöllei,
  Augustinack, Wang, Salat, Ehrlich, Behrens, Jbabdi, Gollub and
  Fischl}]{yendiki_automated_2011}
\bibinfo{author}{Yendiki, A.}, \bibinfo{author}{Panneck, P.},
  \bibinfo{author}{Srinivasan, P.}, \bibinfo{author}{Stevens, A.},
  \bibinfo{author}{Zöllei, L.}, \bibinfo{author}{Augustinack, J.},
  \bibinfo{author}{Wang, R.}, \bibinfo{author}{Salat, D.},
  \bibinfo{author}{Ehrlich, S.}, \bibinfo{author}{Behrens, T.},
  \bibinfo{author}{Jbabdi, S.}, \bibinfo{author}{Gollub, R.},
  \bibinfo{author}{Fischl, B.}, \bibinfo{year}{2011}.
\newblock \bibinfo{title}{Automated {Probabilistic} {Reconstruction} of
  {White}-{Matter} {Pathways} in {Health} and {Disease} {Using} an {Atlas} of
  the {Underlying} {Anatomy}}.
\newblock \bibinfo{journal}{Frontiers in Neuroinformatics} \bibinfo{volume}{5}.
\newblock \URLprefix
  \url{https://www.ncbi.nlm.nih.gov/pmc/articles/PMC3193073/}.
\bibitem[{Çetin et~al.(2014)Çetin, Christensen, Abbott, Stephen, Mayer,
  Cañive, Bustillo, Pearlson and Calhoun}]{cetin_thalamus_2014}
\bibinfo{author}{Çetin, M.S.}, \bibinfo{author}{Christensen, F.},
  \bibinfo{author}{Abbott, C.C.}, \bibinfo{author}{Stephen, J.M.},
  \bibinfo{author}{Mayer, A.R.}, \bibinfo{author}{Cañive, J.M.},
  \bibinfo{author}{Bustillo, J.R.}, \bibinfo{author}{Pearlson, G.D.},
  \bibinfo{author}{Calhoun, V.D.}, \bibinfo{year}{2014}.
\newblock \bibinfo{title}{Thalamus and posterior temporal lobe show greater
  inter-network connectivity at rest and across sensory paradigms in
  schizophrenia}.
\newblock \bibinfo{journal}{NeuroImage} \bibinfo{volume}{97},
  \bibinfo{pages}{117--126}.
\newblock \DOIprefix\doi{10.1016/j.neuroimage.2014.04.009}.
\bibitem[{Çiçek et~al.(2016)Çiçek, Abdulkadir, Lienkamp, Brox and
  Ronneberger}]{cicek_3d_2016}
\bibinfo{author}{Çiçek, O.}, \bibinfo{author}{Abdulkadir, A.},
  \bibinfo{author}{Lienkamp, S.S.}, \bibinfo{author}{Brox, T.},
  \bibinfo{author}{Ronneberger, O.}, \bibinfo{year}{2016}.
\newblock \bibinfo{title}{3d {U}-{Net}: {Learning} {Dense} {Volumetric}
  {Segmentation} from {Sparse} {Annotation}}, in: \bibinfo{booktitle}{Medical
  {Image} {Computing} and {Computer}-{Assisted} {Intervention} – {MICCAI}
  2016}, \bibinfo{publisher}{Springer, Cham}. pp. \bibinfo{pages}{424--432}.
\newblock \URLprefix
  \url{https://link.springer.com/chapter/10.1007/978-3-319-46723-8_49},
  \DOIprefix\doi{10.1007/978-3-319-46723-8_49}.

\end{thebibliography}
\end{document}